\documentclass[lettersize ,journal]{IEEEtran}
\usepackage{amsmath,amsfonts}
\usepackage{algorithmic}
\usepackage{algorithm}
\usepackage{array}
\usepackage[caption=false,font=normalsize,labelfont=sf,textfont=sf]{subfig}
\usepackage{textcomp}
\usepackage{stfloats}
\usepackage{url}
\usepackage{verbatim}
\usepackage{graphicx}
\usepackage{cite}
\usepackage{multirow}
\usepackage{colortbl}
\usepackage{float}
\usepackage{wrapfig}
\usepackage[table]{xcolor} 
\usepackage{colortbl}      
\usepackage{booktabs}
\newcommand{\CompTwo}[2]{%
  \multicolumn{3}{c}{%
    \begin{tabular}{p{0.14\linewidth}p{0.14\linewidth}}
      \centering #1 & \centering #2
    \end{tabular}%
  }%
}

\newcolumntype{C}[1]{>{\centering\arraybackslash}m{#1}}

\begin{document}

\title{Segment Anything with Motion, Geometry, and Semantic Adaptation for Complex Nonlinear Visual Object Tracking}

\author{
Deyi Zhu,
Yuji Wang,
Yong Liu,
Yansong Tang,~\IEEEmembership{Member,~IEEE,}
Bingyao Yu,\\
Jiwen Lu,~\IEEEmembership{Fellow,~IEEE,}
and Jie Zhou,~\IEEEmembership{Fellow,~IEEE,} 

\thanks{The first two authors contribute equally.}
\thanks{Deyi Zhu, Yuji Wang, Yong Liu, and Yansong Tang are with Shenzhen International Graduate School, Tsinghua University, Shenzhen 518055, China (e-mail: zhudy21@mails.tsinghua.edu.cn; tang.yansong@sz.tsinghua.edu.cn).}
\thanks{Bingyao Yu, Jiwen Lu and Jie Zhou are with the Department of Automation, Tsinghua University, Beijing 100084, China.}
\thanks{Yansong Tang is the corresponding author.}}



\maketitle

\begin{abstract}

Traditional visual object tracking (VOT) methods typically rely on task-specific supervised training, limiting their generalization to unseen objects and challenging scenarios with distractors, occlusion, and nonlinear motion. Recent vision foundation models, exemplified by SAM\,2, learn strong video understanding priors from large-scale pretraining and offer a promising foundation for building more robust and generalizable trackers. However, directly applying SAM\,2 to VOT remains suboptimal, as it does not explicitly model target motion dynamics or enforce geometric and semantic consistency across frames, both of which are essential for reliable tracking.
To address this issue, we propose \textbf{SAMOSA}, a new tracking framework that adapts SAM\,2 to complex VOT scenarios by explicitly leveraging motion, geometry, and semantic cues. Specifically, we introduce a lightweight nonlinear motion predictor to model target dynamics and guide mask selection as well as memory filtering. We further exploit semantic cues to detect target shifts and recover from tracking failures, while geometric cues are incorporated as structural constraints to improve tracking stability. In this way, SAMOSA bridges the gap between the implicit video understanding prior of SAM\,2 and explicit tracking-oriented modeling.
Extensive experiments show that SAMOSA consistently outperforms state-of-the-art SAM\,2--based approaches on general benchmarks, demonstrates stronger generalization than supervised VOT methods, and achieves substantial gains on anti-UAV datasets, which typify complex nonlinear motion scenarios. Our code is available at \url{https://github.com/DurYi/SAMOSA}.

\end{abstract}


\section{Introduction}
\label{sec:intro}

\IEEEPARstart{V}{isual} object tracking (VOT) aims to continuously localize a target in a video given its initial state in the first frame. 
Over the past decades, VOT has achieved remarkable progress through Siamese-based trackers~\cite{siamfc, siamrpnpp, siamrcnn, siamon, siamthn}, transformer-based architectures~\cite{transt, ostrack, artrack, artrackv2, bittracker, avtrack}, and large-scale training strategies~\cite{got10k, trackingnet, lasot, otb}. 
Related research has also extended VOT to multimodal settings such as RGB-T tracking~\cite{lsar, tgtrack, siamcda, mambavt}. 
Despite these advances, most existing trackers still rely on task-specific supervised training, which limits their generalization to unseen objects and environments. 
Meanwhile, vision foundation models~\cite{clip, dino, dinov2, dinov3, sam, sam2, sam3} have recently demonstrated strong generalization capabilities across diverse visual tasks. 
However, foundation models specifically designed for visual object tracking remain largely unexplored. 
This motivates the exploration of adapting vision foundation models to VOT in order to build trackers with stronger generalization ability.

\begin{figure}[t]
	\centering
	\includegraphics[width=0.47\textwidth]{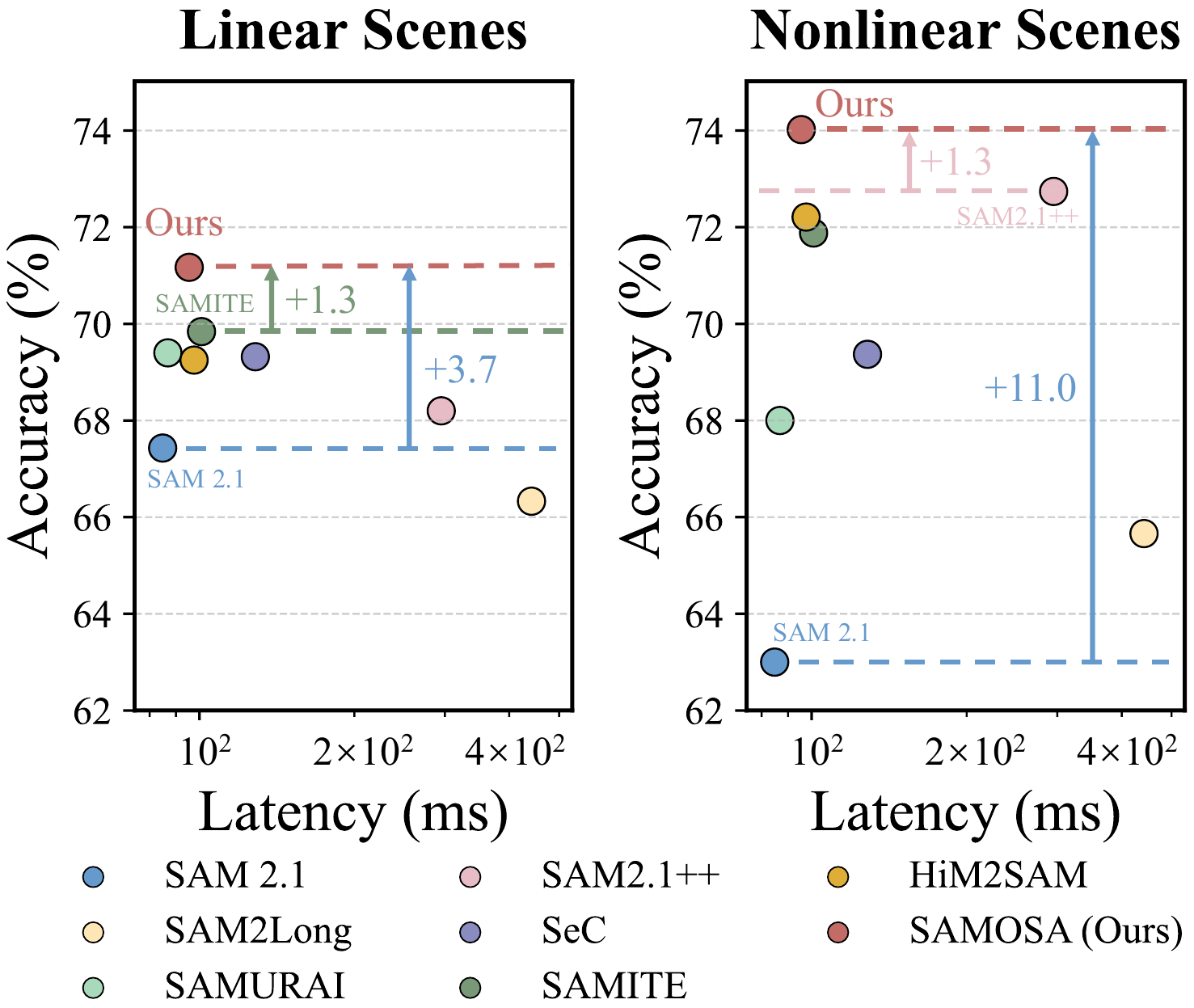}
	\caption{Performance comparison on linear and nonlinear motion scenarios in Anti-UAV300~\cite{antiuav300}. 
    We categorize sequences into \textit{linear} and \textit{nonlinear} splits (see Sec.~\ref{sec:linearnonlinear}). 
    Our method significantly improves SAM\,2's tracking performance in nonlinear scenes with limited latency overhead.}
	\label{fig1}
\end{figure}

\IEEEpubidadjcol

Among recent foundation models, the Segment Anything Model (SAM)~\cite{sam} achieves remarkable success in promptable image segmentation~\cite{scan, ovsnet, scclip} across varying objects. 
Its extension, SAM\,2~\cite{sam2}, generalizes this capability to video object segmentation (VOS)~\cite{vos1, vdn, ravos, lavt1}. 
Benefiting from large-scale pretraining, SAM\,2 demonstrates strong video understanding capability and has been extended to multiple downstream tasks, including visual object tracking~\cite{samurai, dam4sam, samite, him2sam}, camouflage image segmentation~\cite{camopp, sampm, zoomnext}, and audio-visual segmentation~\cite{wang2025sam2, ddavs, diffusiionavs}. 
More recently, SAM\,3~\cite{sam3} further extends SAM models to referring video segmentation~\cite{aamn, vos2, lavt2, iterprime, vgrefiner}.

In general scenarios, existing SAM\,2-based VOT methods achieve strong performance with notable robustness, thanks to carefully designed mask selection and memory management mechanisms~\cite{samurai, dam4sam, sam2long, him2sam, samite, sec}. 
However, they still struggle when targets exhibit complex motion patterns, since they fail to explicitly model nonlinear motion dynamics and to efficiently enforce geometric and semantic consistency during tracking.

In this work, we focus on the challenge of \textit{nonlinear motion}. 
We define \textit{linear motion} as motion that approximately follows constant velocity and smooth displacement across frames, which can be well approximated by constant-velocity models such as the Kalman Filter~\cite{kalmanfilter}. 
In contrast, \textit{nonlinear motion} refers to motion involving velocity variations, such as acceleration, direction changes, camera movements, shape variations, or temporary disappearance of the target. 
Such nonlinear dynamics frequently occur in real-world VOT scenarios, significantly increasing tracking difficulty and cannot be well approximated by constant-velocity models, thus requiring motion models capable of capturing nonlinear dynamics.

\begin{figure*}[t]
	\centering	\includegraphics[width=\textwidth]{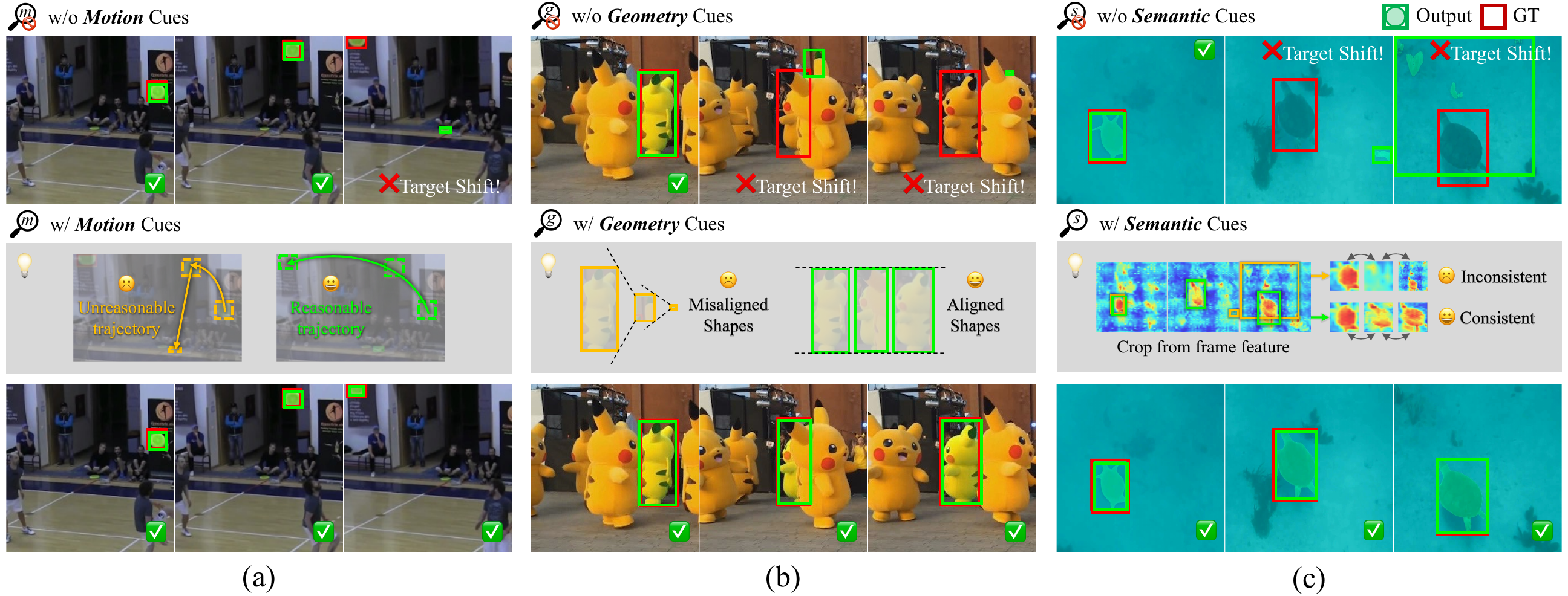}
	\caption{Examples of the roles of \textit{motion}, \textit{geometry}, and \textit{semantic} cues in complex visual object tracking. Frames are cropped for clarity. 
    (a) \textit{Motion} cues help track small objects moving in cluttered backgrounds. 
    (b) \textit{Geometry} cues help prevent interference from similar distractors nearby. 
    (c) \textit{Semantic} cues utilize latent feature to help identify and prevent target shift errors.}
	\label{fig2}
\end{figure*}

To better address these challenges, we observe that effective visual object tracking fundamentally relies on three complementary cues: \textit{motion}, \textit{geometry}, and \textit{semantics}. 
As illustrated in Figure~\ref{fig2}, \textit{motion} describes the temporal evolution of the target position and provides predictive dynamics for associating objects across frames.
\textit{Geometry} captures intrinsic low-level visual properties such as shape, area, and boundary structure, helping distinguish the target from distractors. 
\textit{Semantics} encodes high-level appearance and contextual information, ensuring consistent identification of the target despite viewpoint or illumination changes. 
A robust tracker should therefore integrate these cues to jointly model temporal coherence, spatial stability, and semantic consistency.
Although SAM\,2 implicitly captures aspects of these cues through large-scale pretraining, it lacks explicit modeling and constraint mechanisms, making it prone to tracking failures in complex scenarios shown in Figure~\ref{fig2}. 
Several existing SAM\,2-based tracking methods partially exploit one or two of these cues, but their designs remain coarse and limited. For example, simple motion prediction strategies~\cite{samurai, him2sam} may fail in scenes with nonlinear motion dynamics. The exploiting of \textit{semantics}~\cite{samite} also did not serve for explicitly detect tracking failures. Besides, none of existing methods fully integrate all three cues in a unified framework.

Based on this insight, we propose a new tracking framework, \textbf{SAMOSA} (\textbf{S}egment \textbf{A}nything with \textbf{M}otion, Ge\textbf{O}metry, and \textbf{S}emantic \textbf{A}daptation), designed for complex nonlinear motion scenarios. 
To explicitly model motion dynamics, we introduce a Motion Predictor (MP) based on a higher-order Markov model that captures nonlinear target motion patterns. 
The predicted \textit{motion} and \textit{geometry} cues are used to guide mask selection, enabling more reliable temporal associations. 
We further develop an Error Detection–Recovery Module (EDRM) that detects potential tracking failures during inference and triggers recovery using \textit{geometry} and \textit{semantic} cues. 
Moreover, we propose a Target-Aware Memory Bank (TAMB) that integrates mask quality, target visibility, and \textit{motion} information to prioritize reliable memory frames.

Notably, MP is the only trainable component in our framework. 
It is trained solely on annotated bounding-box trajectories without relying on video frames and can be seamlessly integrated into SAM\,2 during inference.

We evaluate our method on multiple VOT benchmarks including LaSOT$_{ext}$~\cite{lasotext}, OTB~\cite{otb}, TrackingNet~\cite{trackingnet}, and Anti-UAV series~\cite{antiuav300, antiuav410, antiuav600, dutdatset}. 
Experimental results demonstrate that SAMOSA consistently outperforms existing trackers with stronger generalization ability and achieves substantial improvements on challenging nonlinear motion scenarios.

Our main contributions are summarized as follows:
\begin{itemize}
    \item We propose a higher-order Markov motion predictor to model nonlinear motion, together with an error detection–recovery module that explicitly identifies potential tracking failures and mitigates error propagation.
    \item We develop a target-aware memory bank that adaptively selects representative and reliable memory frames guided by confidence, occlusion, and motion cues.
    \item Our method achieves state-of-the-art performance on general VOT benchmarks and challenging anti-UAV tracking benchmarks, outperforming previous approaches.
\end{itemize}

\section{Related Work}
\label{sec:related_work}

\subsection{Conventional Visual Object Tracking}

Visual object tracking (VOT) has evolved significantly over the past decade. Early trackers~\cite{mosse, kcf, csrdcf} rely on correlation filters for efficient tracking. With the rise of deep learning, Siamese-network-based methods such as SiamFC~\cite{siamfc} and SiamRPN++~\cite{siamrpnpp} formulate tracking as similarity learning between template and search regions. Another line of work explores online discriminative learning, where DiMP~\cite{dimp} learns a target-specific classifier to handle appearance variations. Recent progress is largely driven by transformer-based architectures and end-to-end modeling. TransT~\cite{transt} introduces attention-based feature fusion, while OSTrack~\cite{ostrack} proposes a unified one-stream framework for holistic feature interaction. More recent works, including LoRAT~\cite{lorat}, ODTrack~\cite{odtrack}, and ARTrackV2~\cite{artrackv2}, further improve robustness through efficient adaptation and temporal modeling.

Despite these advances, existing trackers still struggle with long-term occlusion, rapid appearance variation, complex nonlinear motion, and generalization to unseen targets and environments. A key reason is that most existing trackers rely on task-specific supervised training, which restricts cross-domain generalization. Foundation models such as SAM\,2~\cite{sam2}, however, demonstrate strong adaptability to unseen domains, highlighting their potential for visual object tracking tasks.

\subsection{Video Object Segmentation for Visual Object Tracking}

Video object segmentation (VOS) naturally suits tracking non-rigid or irregularly shaped objects. Compared to bounding boxes, segmentation masks can adapt to complex contours and structural variations, enabling robust tracking. Recent foundation models for VOS, such as SAM\,2~\cite{sam2}, exhibit strong zero-shot segmentation and tracking capabilities. However, they still struggle in scenarios involving occlusions, distractors, or multiple similar objects. Recent studies address these issues mainly from memory management and motion modeling. In terms of memory management, SAM2Long~\cite{sam2long} constructs a constrained tree memory structure for long-term and ambiguous cases, at the cost of higher computation. SAM2.1++~\cite{dam4sam} and HiM2SAM~\cite{him2sam} design long-short memory hierarchies to enhance robustness and temporal consistency, while SeC~\cite{sec} expands the temporal window of the memory bank. SAMITE~\cite{samite} selects memory entries using feature- and position-wise anchors, all aiming to refine the FIFO memory policy. For motion modeling, SAMURAI~\cite{samurai} integrates a Kalman Filter (KF) to mitigate ambiguous predictions. However, under the constant-velocity assumption, the linear KF struggles to capture nonlinear dynamics. HiM2SAM~\cite{him2sam} introduces point trackers for complex scenarios but still fails to capture consistent motion trends. 

Despite recent progress, existing methods still struggle in nonlinear scenes, as illustrated in Figure~\ref{fig1}, and no approach effectively adapts SAM\,2 to handle such dynamics without substantial computational cost. To address this gap, we introduce \textbf{SAMOSA}, a lightweight enhancement of SAM\,2 for complex nonlinear visual object tracking.

\begin{figure*}[t]
	\centering	\includegraphics[width=\textwidth]{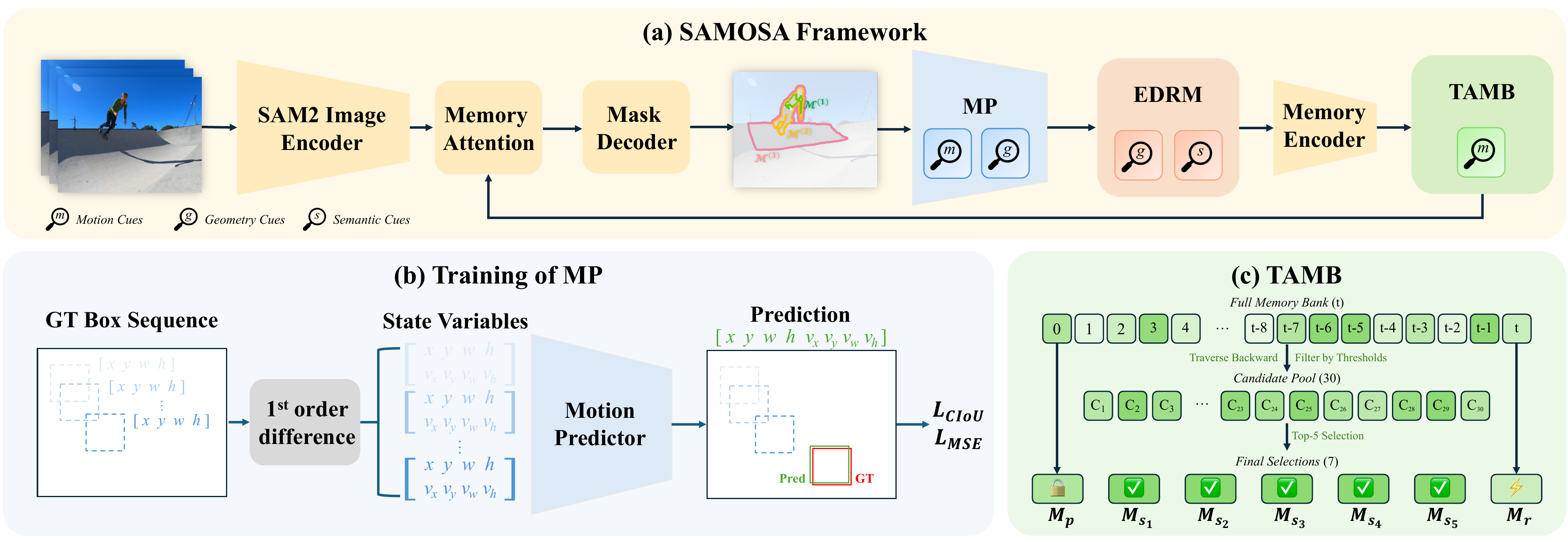}
	\caption{(a) Overall pipeline of \textbf{SAMOSA}, which integrates the proposed MP, EDRM, and TAMB modules into the SAM\,2 backbone.
        (b) The MP is trained independently from SAM\,2 and videos. After training, it is directly plugged into SAM\,2 for inference.
        (c) The framework of TAMB, consisting of a memory filtering stage and a top-$k$ selection process.}
	\label{fig3}
\end{figure*}

\section{Preliminary}
\label{sec:preliminary}

SAM\,2 employs the pre-trained Hiera~\cite{hiera} as a vision encoder to extract features from each frame. These features are refined through memory-attention with historical representations stored in a memory bank. The memory-conditioned features are decoded into $N=3$ candidate masks $\{\mathcal{M}^{(i)}\}_{i=1}^N$ by a bidirectional transformer, while two MLP heads predict the corresponding IoU ($S_{IoU}$) and object ($S_{obj}$) scores. Here, $S_{IoU}$ measures mask affinity and quality, and $S_{obj}$ estimates the target’s visibility. The decoded masks are further processed and inserted into the memory bank via a FIFO queue, preserving spatial and semantic information of tracked objects. 

Despite its strong generalization to diverse visual domains, directly applying SAM\,2 to complex nonlinear tracking scenarios remains challenging. It lacks explicit motion modeling of historical trajectories and selects masks solely according to $S_{IoU}$, which is inadequate for tasks requiring more comprehensive decision criteria. Robust tracking instead demands integrated consideration of \textit{motion}, \textit{geometry} and \textit{semantic} cues to ensure consistent object localization across time.

\section{Method}
\label{sec:method}

The overall pipeline of our proposed \textbf{SAMOSA} is illustrated in Figure~\ref{fig3}(a). It integrates essential cues into SAM\,2’s mask selection and memory attention mechanisms to better handle the non-linear dynamics of targets. The Motion Predictor (MP) predominates under stable conditions, leveraging \textit{motion} and \textit{geometry} cues to guide mask selection, while the Error Detection-Recovery Module (EDRM) serves as a safeguard that overrides it in uncertain situations by exploiting \textit{geometry} and \textit{semantic} cues to detect and rectify errors. This hybrid design ensures robustness against both gradual motion patterns and abrupt changes in motion dynamics. Meanwhile, the Target-Aware Memory Bank (TAMB) leverages \textit{motion} cues to perform filtering and selection over memory frames, yielding temporally consistent and high-quality historical priors that further enhance mask generation.

\subsection{Motion Predictor (MP)}
\label{subsec:mp}
\textbf{Non-linear Motion Prediction.} The previous linear predictor~\cite{samurai}, built under constant-velocity and first-order Markov assumptions, defines a state transition matrix $F$ to predict the next state ${\boldsymbol{\hat s}_{t+1}}$ from the previous ${\boldsymbol{\hat s}_t}$. This process can be formally expressed in Equation~(\ref{eq2}):
\begin{subequations}\label{eq2}  
  \begin{align}  
    \boldsymbol{s} &= {[x,y,w,h,\dot x,\dot y,\dot w,\dot h]^T} , \\
    {\boldsymbol{\hat s}_{t + 1}} &= F{\boldsymbol{\hat s}_t} ,
  \end{align}  
\end{subequations}
where $\boldsymbol{s}$ denotes the bounding-box state vector, including the center coordinates $(x, y)$, width $w$, height $h$, and their first-order derivatives indicated by the dot notation. This strategy works well when the target follows constant-velocity and straight-line motion patterns. However, VOT tasks often exhibit short-term temporal coherence with non-linear dynamics. The speed and direction of motion is usually not fixed. Such a simplification struggles to capture complex non-linear motion patterns in these scenarios. 

To address this limitation, we introduce a Motion Predictor (MP), a sequence model based on a $k$-th order Markov framework, where the prediction at time $t{+}1$ is conditioned on a sliding window of the past $k$ states:
\begin{equation}
\boldsymbol{\hat{s}}_{t+1} = f_\theta(\boldsymbol{\widetilde s}_t, \boldsymbol{\widetilde s}_{t-1}, \dots, \boldsymbol{\widetilde s}_{t-k+1}), 
\end{equation}
where $f_\theta$ parameterizes the non-linear state transition, and $\boldsymbol{\widetilde s}_t, \boldsymbol{\widetilde s}_{t-1}, \dots, \boldsymbol{\widetilde s}_{t-k+1}$ denotes the measurement states derived from previously selected SAM\,2 masks during inference, or ground-truth states during training. Unlike models that require access to the entire sequence, this design extends the Markov assumption to a finite history, effectively balancing modeling capacity and computational efficiency.

\textbf{Training of MP}. MP can be trained independently using only annotated bounding-box trajectories available in standard VOT benchmarks, as in Figure~\ref{fig3}(b). We leverage the mean squared error (MSE) and complete IoU (CIoU) \cite{ciou_loss} loss for supervision. The CIoU loss improves overlap area consistency, reduces center point displacement, and ensures better alignment of the aspect ratio between the predicted bounding box and the ground truth. The overall regression loss is defined as:
\begin{equation}
    \mathcal{L}_{\text{reg}} = \lambda_1\mathcal{L}_{\text{MSE}} + \lambda_2\mathcal{L}_{\text{CIoU}} ,
\end{equation}
where $\lambda_1$ and $\lambda_2$ are the corresponding loss weights. After training, the MP functions as a plug-and-play module that can be seamlessly integrated into SAM\,2 for inference. 

\textbf{Mask Selection.} 
During inference, we maintain a FIFO history state bank that stores the most recent $k$ outputs. 
At each time step, when the mask decoder generates the mask for the current frame, the MP also predicts a bounding box $\mathcal{B}_\mathrm{MP}$ based on the stored historical bounding boxes, which is subsequently used to guide mask selection.


We integrate \textit{geometry} and \textit{motion} cues into the mask selection process by introducing a geometric score $S^{(n)}_g$ and a motion score $S^{(n)}_m$ for each mask $\mathcal{M}^{(n)}$. For accurate tracking, the predicted box $\mathcal{B}_\mathrm{MP}$ and the $N=3$ candidate boxes $\{\mathcal{B}^{(n)}\}_{n=1}^N$ derived from $\{\mathcal{M}^{(n)}\}_{n=1}^N$ should remain consistent in shape, scale and spatial position. Accordingly, we define the geometric score $S_g$ as a weighted combination of (1) the similarity of the aspect ratio (AR) and (2) the similarity of the area between $\mathcal{B}_\mathrm{MP}$ and $\mathcal{B}^{(n)}$, capturing their geometric consistency. Meanwhile, the motion score $S_m$ is computed as the IoU between $\mathcal{B}_\mathrm{MP}$ and $\mathcal{B}^{(n)}$, which measures spatial alignment with the motion-predicted trajectory. Thus, the geometric score and motion score are defined as follows:
\begin{subequations}\label{eq5new}  
  \begin{align}  
    S^{(n)}_{\text{AR}} &= 
    \mathrm{Sim}(\mathrm{AR}(\mathcal{B}_\mathrm{MP}), \mathrm{AR}(\mathcal{B}^{(n)})),\\ 
    S^{(n)}_{\text{Area}} &= 
    \mathrm{Sim}(\mathrm{Area}(\mathcal{B}_\mathrm{MP}), \mathrm{Area}(\mathcal{B}^{(n)})),\\
    S^{(n)}_g &= \beta_{\text{AR}}S^{(n)}_{\text{AR}}+\beta_{\text{Area}}S^{(n)}_{\text{Area}},\\
    S^{(n)}_m &= \mathrm{IoU}(\mathcal{B}_\mathrm{MP}, \mathcal{B}^{(n)}),
  \end{align}  
\end{subequations}
where $\mathrm{Sim}(x,y)={\min(x, y)}/{\max(x, y)}$, and $\mathrm{IoU}(\cdot,\cdot)$ denotes the Intersection-over-Union (IoU) between two boxes.

Different from SAM\,2, which selects the output mask solely based on the IoU score $S_{IoU}$, we further incorporates $S_g$ and $S_m$ to evaluate each candidate. The final mask is then selected according to the highest weighted combination of the three scores, as formulated in Equation~(\ref{eq5}):
\begin{equation}
    \mathcal{M}^* = \arg\max_{\mathcal{M}^{(n)}} \left( 
        \alpha S^{(n)}_{IoU} 
        + S^{(n)}_g 
        + \gamma S^{(n)}_m
    \right),
    \label{eq5}
\end{equation}
where $\alpha$ and $\gamma$ are their corresponding weights. With the assistance of MP, the selected mask not only exhibits high affinity with the target, but also conforms to the physical motion patterns, thereby enhancing tracking robustness.



\subsection{Error Detection-Recovery Module (EDRM)}
\begin{figure}[t]
	\centering	\includegraphics[width=0.47\textwidth]{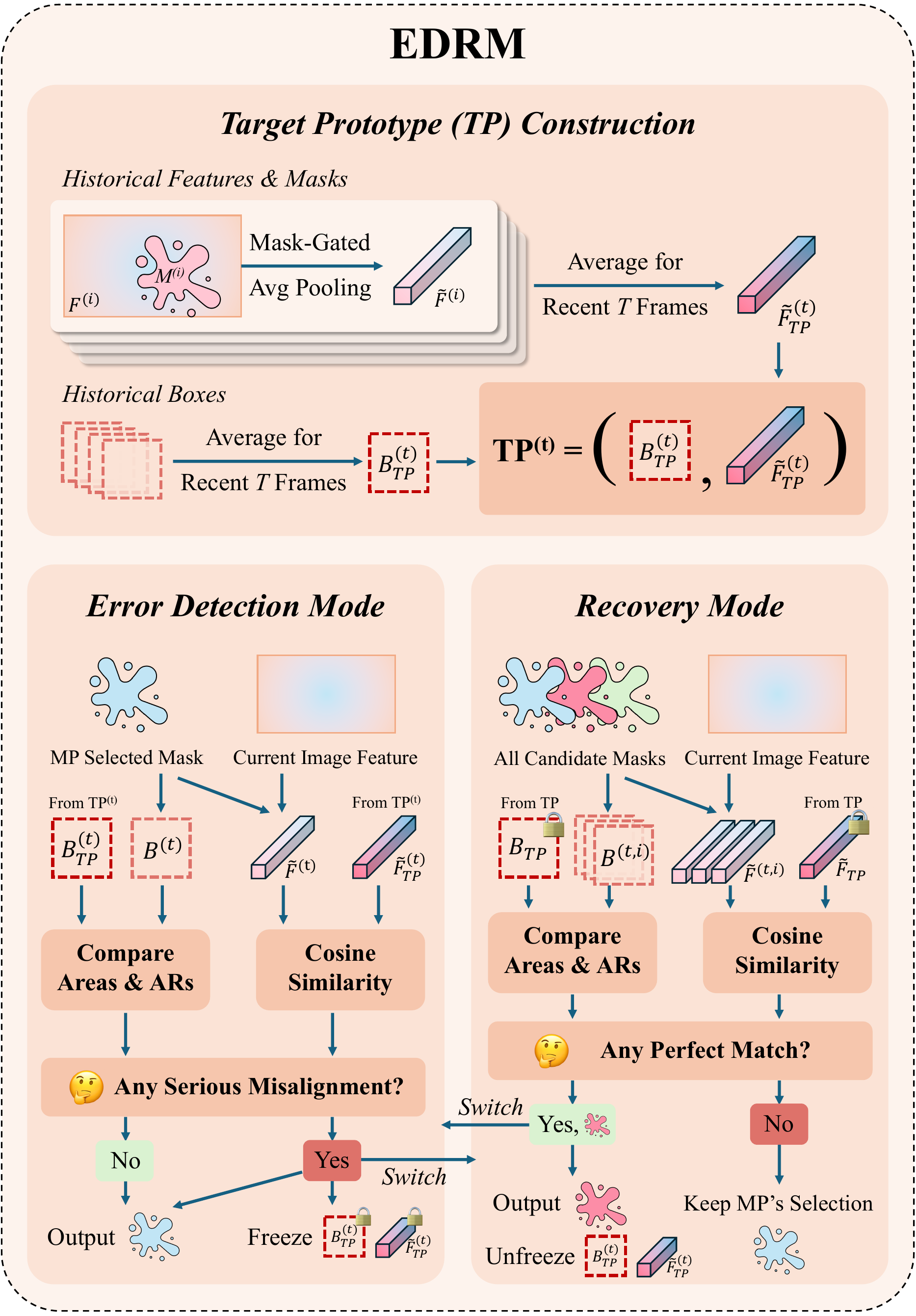}
	\caption{The framework of Error Detection-Recovery Module.}
	\label{fig4}
\end{figure}

Even with the assistance of MP, tracking errors may still arise due to factors such as camera shake, target occlusion, or nearby distractors. To mitigate the risk of error accumulation, we introduce Error Detection–Recovery Module (EDRM) as shown in Figure~\ref{fig4}, designed to detect and recover from tracking failures. EDRM is built upon the assumption that the target’s visual state remains relatively stable over short temporal intervals. Accordingly, it maintains a Target Prototype (TP) that represents the recent reliable states of the target, and identifies tracking errors by measuring the \textit{geometric} and \textit{semantic} misalignment between the current output and TP.

\textbf{Target Prototype (TP).} During inference, the TP is constructed using the outputs from the most recent $T$ frames to capture both \textit{geometry} and \textit{semantic} cues. 
For \textit{geometry} cues, we average the bounding boxes from the latest $T$ outputs before time step $t$ to obtain the geometric representation $\mathcal{B}_{\mathrm{TP}}^{(t)}$. 
For \textit{semantic} cues, we leverage the image embeddings encoded by SAM\,2's Hiera~\cite{hiera} encoder. 
Given the image embeddings $F^{(i)} \in \mathbb{R}^{h \times w \times d}$ and the corresponding mask $\mathcal{M}^{(i)}$ of the $i$-th frame, we apply mask-gated average pooling on $F^{(i)}$ to obtain $\tilde{F}^{(i)}$, a compact semantic representation of the target in frame $i$, as shown in Equation~(\ref{eq6}a). 
At time step $t$, $\tilde{F}_{\mathrm{TP}}^{(t)}$ is computed by averaging $\tilde{F}^{(i)}$ over the most recent $T$ frames, as shown in Equation~(\ref{eq6}b). 
Finally, the TP at time step $t$ is represented as the combination of $\mathcal{B}_{\mathrm{TP}}^{(t)}$ and $\tilde{F}_{\mathrm{TP}}^{(t)}$, as defined in Equation~(\ref{eq6}c).
\begin{subequations}\label{eq6}  
  \begin{align}  
    \tilde{F}^{(i)} &= 
    \frac{\sum_x \sum_y F^{(i)}(x,y) \cdot \mathbb{I}\!\left[\mathcal{M}^{(i)}(x,y)=1\right]}
    {\sum_x \sum_y \mathbb{I}\!\left[\mathcal{M}^{(i)}(x,y)=1\right]},\\
    \tilde{F}_{\mathrm{TP}}^{(t)}&=\frac{1}{T}\sum_{i=t-T}^{t-1} \tilde{F}^{(i)}, \\
    {\mathrm{TP}}^{(t)} &= \left(\mathcal{B}_{\mathrm{TP}}^{(t)},\,\tilde{F}_{\mathrm{TP}}^{(t)} \right).
  \end{align}  
\end{subequations}
The TP is updated throughout tracking until the EDRM detects a potential error, upon which the TP is temporarily frozen to avoid contamination from erroneous outputs.

\textbf{Error Detection and Recovery.} 
EDRM is inserted after the mask selection process and is initialized in the error detection mode. 
At each time step $t$, it uses the image embeddings $F^{(t)}$ and output mask $\mathcal{M}^{(t)}$ of the current frame to obtain the bounding box $\mathcal{B}^{(t)}$ and compact semantic representation $\tilde{F}^{(t)}$ (similar to Equation~(\ref{eq6}a)), which are then compared with $\mathcal{B}_{\mathrm{TP}}^{(t)}$ and $\tilde{F}_{\mathrm{TP}}^{(t)}$ from TP to evaluate their similarity in aspect ratio (AR), area, and semantics. Thus, we obtain three similarity scores, as formulated in Equation~(\ref{eq7}):
\begin{subequations}\label{eq7}  
  \begin{align}  
    S^{(t)}_{ar} &= 
    \mathrm{Sim}(\mathrm{AR}(\mathcal{B}^{(t)}), \mathrm{AR}(\mathcal{B}_{\mathrm{TP}}^{(t)})),\\
    S^{(t)}_a &= 
    \mathrm{Sim}(\mathrm{Area}(\mathcal{B}^{(t)}), \mathrm{Area}(\mathcal{B}_{\mathrm{TP}}^{(t)})),\\
    S^{(t)}_s &= 
    \mathrm{CosSim}(\tilde{F}^{(t)}, \tilde{F}_{\mathrm{TP}}^{(t)}),
  \end{align}  
\end{subequations}
where $\mathrm{Sim}(x,y)={\min(x, y)}/{\max(x, y)}$, and $\mathrm{CosSim}$ refers to cosine similarity.
If any of the three scores drops below its corresponding predefined threshold 
$\sigma_{ar}$, $\sigma_{a}$, or $\sigma_{s}$, 
EDRM flags a potential tracking error and switches to recovery mode.

Once entering the recovery mode, TP is frozen, while MP continues to select masks based on its predictions. During this phase, EDRM changes its role to actively seeking an opportunity to correct the tracking error. At each time step $t$, it utilizes TP and all $N=3$ candidate masks to compute $\{ S^{(t,n)}_{ar}, S^{(t,n)}_a, S^{(t,n)}_s\}_{n=1}^{N}$. If there exists a candidate whose scores all exceed the predefined thresholds $\tau_{ar}$, $\tau_{a}$, and $\tau_{s}$, it is regarded as the correct target with high confidence. EDRM then overwrites MP’s choice with this candidate, resumes TP updating, and switches back to the error detection mode.

\subsection{Target-Aware Memory Bank (TAMB)}

Memory selection is also crucial for motion modeling, as it ensures the prerequisite generation of high-quality masks, while low-quality masks may propagate errors to subsequent predictions. To address this, we propose TAMB, a target-aware memory bank that utilizes a threshold-based top-$k$ selection strategy for memory management, as illustrated in Figure~\ref{fig3}(c). TAMB selects memory frames containing the most representative information of the target, based on three complementary elements: \textit{motion} cues, mask quality, and target completeness.

For any frame $i$, SAM\,2's mask decoder outputs an IoU score $S^{(i)}_{IoU}$ and an object score $S^{(i)}_{obj}$, which respectively indicate (1) the quality of the predicted mask and (2) the likelihood that the target is visible in the frame without occlusion.
These scores are utilized to identify memory frames with reliable segmentation quality and clear target appearance. In addition, the motion score $S^{(i)}_m$ from MP serves as a \textit{motion} cue, helping to identify frames with stable target motion and filter out those violating regular motion patterns.

First, to preserve short-term temporal information, we always retain the most recent memory frame, denoted as $M_r$. 
Then, leveraging $S^{(i)}_{IoU}$, $\mathrm{sigmoid}(S^{(i)}_{obj})$, and $S^{(i)}_m$, 
we traverse backward in time to collect frames that meet the predefined thresholds 
$\mu_{IoU}$, $\mu_{obj}$, and $\mu_{m}$ until $M=30$ candidate memory frames are obtained. 
For each candidate frame, a weighted sum of the three scores is computed as in Equation~(\ref{eq8}), 
and following SAMITE~\cite{samite}, the top $(N_m - 1)$ frames with the highest weighted scores are selected, where $N_m = 6$ represents the maximum number of unprompted memory slots. 
Finally, consistent with SAM\,2, the memory $M_p$ from the prompted frame is always retained to provide the initial condition for tracking.
\begin{equation}
    S_\mathrm{TAMB}^{(i)} = 
        \delta S^{(i)}_{IoU} 
        + \epsilon \, \mathrm{sigmoid}(S^{(i)}_{obj})
        + \zeta S^{(i)}_m,
    \label{eq8}
\end{equation}
where $\delta$, $\epsilon$ and $\zeta$ are weighting coefficients. We apply $\mathrm{sigmoid}(\cdot)$ on $S^{(i)}_{obj}$ to map its value into the range $[0, 1]$.

\begin{table*}[t]
    \centering
    \caption{Comparison of performance and inference speed with supervised VOT methods and SAM\,2-based methods on general-purpose VOT benchmarks. All $\Delta\text{Latency}$ values are tested on $\text{LaSOT}_{ext}$ and reported relative to SAM\,2.1.}
    \resizebox{\linewidth}{!}{
    \begin{tabular}{lccccccccccc}
        \toprule
        \multirow{2}{*}{Methods} & \multicolumn{3}{c}{$\text{LaSOT}_{ext}$ (\%)} & \multicolumn{3}{c}{OTB (\%)} & \multicolumn{3}{c}{TrackingNet (\%)} & \multirow{2}{*}{$\Delta\text{Latency}$ (ms)} & \multirow{2}{*}{VENUE} \\
        \cmidrule(lr){2-4} \cmidrule(lr){5-7} \cmidrule(lr){8-10}
         & ${AUC}$ & ${P}$ & ${P_{norm}}$ & ${AUC}$ & ${P}$ & ${P_{norm}}$ & ${Succ}$ & ${P}$ & ${P_{norm}}$ \\
        \midrule 
        \multicolumn{12}{c}{\textit{Supervised VOT methods}} \\
        \midrule 
        $\text{OSTrack}_{384}$~\cite{ostrack} & 50.5 & 57.6 & 61.3 & 55.9 & - & - & 83.9 & 83.2 & 88.5 & - & ECCV'2022 \\
        $\text{SeqTrack-L}_{384}$~\cite{seqtrack} & 50.7 & 57.5 & 61.6 & - & - & - & 85.5 & 85.8 & 89.8 & - & CVPR'2023 \\
        $\text{ROMTrack}_{384}$~\cite{romtrack} & 51.3 & 58.6 & 62.4 & 70.9 & - & - & 84.1 & 83.7 & 89.0 & - & ICCV'2023 \\
        HIPTrack~\cite{hiptrack} & 53.0 & 60.6 & 64.3 & 71.0 & - & - & 84.5 & 83.8 & 89.1 & - & CVPR'2024 \\
        $\text{ARTrackV2-L}_{384}$~\cite{artrackv2} & 53.4 & 60.2 & 63.7 & - & - & - & \textbf{86.1} & 86.2 & 90.4 & - & CVPR'2024 \\
        ODTrack-L~\cite{odtrack} & 53.9 & 61.7 & 65.4 & - & - & - & \textbf{86.1} & 86.7 & 91.0 & - & AAAI'2024\\
        $\text{LoRAT-L}_{378}$~\cite{lorat} & 56.6 & 65.1 & 69.0 & \textbf{72.0} & - & - & 85.6 & 85.4 & 89.7 & - & ECCV'2024 \\
        \midrule 
        \multicolumn{12}{c}{\textit{SAM\,2-based methods}} \\
        \midrule 
        SAM\,2~\cite{sam2} & 56.89 & 66.38 & 69.54 & 69.41 & 91.49 & 82.51 & 84.94 & 87.80 & 90.99 & -0.8 & {ICLR'2025} \\
        SAM\,2.1~\cite{sam2} & 58.19 & 68.08 & 70.57 & 71.61 & 93.92 & 85.36 & 85.41 & \textbf{88.37} & 91.40 & 0 &{ICLR'2025} \\
        SAM2Long ~\cite{sam2long} & 61.17 & 72.46 & 74.21 & 71.91 & \textbf{94.32} & \underline{86.17} & 85.42 & 88.42 & \underline{91.41} & +357.2 & {ICCV'2025} \\
        SAMURAI ~\cite{samurai}& 60.96 & 72.06 & 73.84 & 71.71 & 93.99 & 85.41 & 85.39 & \underline{88.33} & 91.34 & +2.0 &{Arxiv'2024} \\
        SAM2.1++~\cite{dam4sam} & 59.55& 70.57 & 71.98 & 70.82 & 91.65 & 83.29 & 84.78 & 87.76 & 91.25 & +210.3 & {CVPR'2025} \\
        SeC ~\cite{sec}& 60.27 & 71.15 & 72.74 & 70.55 & 91.51 & 83.81 & 85.27 & 88.05 & 91.23 &  +43.6 & {ICLR'2026} \\
        SAMITE ~\cite{samite}& \underline{62.22} & \underline{73.70} & \underline{75.35}  & 71.05 & 92.30 & 84.27 & 85.22 & 88.08 & 91.38 & +16.1 & {Arxiv'2025} \\
        HiM2SAM ~\cite{him2sam}& 62.05 & 73.16 & 74.91 & 71.32 & 92.87 & 84.63 & 85.40 & 88.25 & 91.40 & +12.8 & {PRCV'2025} \\
        \hline
        \rowcolor{gray!20} 
        SAMOSA (Ours) & \textbf{62.97} & \textbf{74.20} & \textbf{75.96} & \underline{71.94} & \underline{94.26} & \textbf{86.44 } & \underline{85.55} & 88.31 & \textbf{91.43} & +10.7 & - \\
        \bottomrule
    \end{tabular}
    }
    \label{tab1}
\end{table*}

\section{Experiments}
\label{sec:experiments}

\subsection{Experimental Settings}
\label{ExperimentalSettings}

\textbf{Datasets.} We evaluate \textbf{SAMOSA} on three general-purpose VOT benchmarks and four anti-UAV tracking benchmarks.

\textit{General-purpose VOT benchmarks.} 
(1) \textbf{$\text{LaSOT}_{ext}$}~\cite{lasotext}, an extended and more challenging version of LaSOT~\cite{lasot}, contains 150 videos with an average of 2,393 frames per video. 
(2) \textbf{OTB}~\cite{otb} contains 100 videos with an average of 598 frames per video. 
(3) \textbf{TrackingNet}~\cite{trackingnet} contains 511 videos with an average of 441 frames per video. 
These datasets serve as representative benchmarks for evaluating trackers on typical challenging scenarios, including object deformation, frequent occlusion, same-class distractors, and high-level semantic reasoning.

\textit{Anti-UAV tracking benchmarks.} 
(1) \textbf{Anti-UAV300}~\cite{antiuav300} contains 91 paired RGB and thermal infrared (TIR) videos, each with an average of 938 frames per video. 
(2) \textbf{Anti-UAV410}~\cite{antiuav410} contains 120 TIR videos with an average of 1,081 frames per video. 
(3) \textbf{Anti-UAV600}~\cite{antiuav600} (validation set) contains 50 TIR videos with an average of 1,179 frames per video. 
(4) \textbf{DUT\,Anti-UAV}~\cite{dutdatset} contains 20 RGB videos with an average of 1,240 frames per video. 
These anti-UAV tracking datasets pose more challenging conditions, featuring highly nonlinear motion patterns, camera shakes, small targets, cross-modal adaptability and sparse semantic cues. 

Together, the two types of benchmarks evaluate trackers across different difficulty levels and provide a comprehensive assessment of model robustness in diverse scenarios.

\textbf{Baselines.} We compare our methods with SAM\,2-based methods~\cite{sam2, sam2long, samurai, dam4sam, sec, samite, him2sam}, 
as well as representative supervised VOT methods~\cite{ostrack, seqtrack, romtrack, hiptrack, artrackv2, odtrack, lorat}.

\textbf{Implementation Details.} In our method, only the MP is trained, while TAMB and EDRM require no training, and all modules of the SAM\,2.1 backbone remain frozen to ensure a fair comparison.
The MP employs a 4-layer LSTM network~\cite{lstm} that predicts the target state based on the past \(k = 5\) frames. It is trained solely on trajectory annotations from LaSOT~\cite{lasot}, and then integrated into our method for evaluation on all other benchmarks to assess its generalization ability. SAM\,2~\cite{sam2} uses the \textit{large}-size checkpoint, while all other SAM\,2-based methods use the \textit{large}-size SAM\,2.1~\cite{sam2} checkpoint. For supervised VOT methods, we report results from those with parameter counts comparable to SAM\,2.1-\textit{large}. All experiments are conducted on a single NVIDIA RTX 3090 GPU with 24\,GB of memory.

\textbf{Evaluation Metrics.} For $\text{LaSOT}_{ext}$~\cite{lasotext} and OTB~\cite{otb}, we adopt the area under the curve (AUC), precision (P), and normalized precision ($\text{P}_{norm}$) following SAMURAI~\cite{samurai}. For TrackingNet~\cite{trackingnet}, we evaluate using success rate (Succ), precision (P), and normalized precision ($\text{P}_{norm}$) on its official evaluation platform. For anti-UAV tracking benchmarks, we adopt average overlap accuracy (Acc)~\cite{antiuav600}, area under the curve (AUC), and precision (P), which are three commonly used metrics for this task.

\begin{table*}[t]
    \centering
    \caption{Comparison of performance with SAM\,2-based methods on representative anti-UAV tracking benchmarks.}
    \resizebox{\linewidth}{!}{
    \begin{tabular}{lccccccccccccccc}
        \toprule
        \multirow{2}{*}{Methods} & \multicolumn{3}{c}{Anti-UAV300 RGB (\%)} & \multicolumn{3}{c}{Anti-UAV300 TIR (\%)} & \multicolumn{3}{c}{Anti-UAV410 (\%)} & \multicolumn{3}{c}{Anti-UAV600 (\%)} & \multicolumn{3}{c}{DUT\,Anti-UAV (\%)}\\
        \cmidrule(lr){2-4} \cmidrule(lr){5-7} \cmidrule(lr){8-10} \cmidrule(lr){11-13} \cmidrule(lr){14-16}
         & ${Acc}$ & ${P}$ & ${AUC}$ & ${Acc}$ & ${P}$ & ${AUC}$ & ${Acc}$ & ${P}$ & ${AUC}$ & ${Acc}$ & ${P}$ & ${AUC}$ & ${Acc}$ & ${P}$ & ${AUC}$ \\
        \midrule 
        SAM\,2~\cite{sam2} & 65.44 & 91.16 & 58.37 & 56.95 & 88.03 & 54.75 & 54.38 & 84.33 & 51.99 & 45.16 & 64.97 & 36.82 & 63.26 & 92.97 & 57.52 \\
        SAM\,2.1~\cite{sam2} & 67.18 & 89.73 & 60.05& 59.32 & 86.86 & 57.04 & 55.93 & 83.04 & 53.51 & 45.38 & 62.89 & 37.66 & 63.84 & 89.35 & 58.01 \\
        SAM2Long ~\cite{sam2long} & 66.28 & 88.38 & 59.00 & 58.51 & 86.06 & 56.16 & 56.68 & 83.99 & 54.08 & 47.53 & 64.16 & 38.61 & 66.91 & 93.55 & 61.11 \\
        SAMURAI ~\cite{samurai}& 69.30 & 92.81 & 62.28 & 61.53 & 90.44 & 59.19 & 58.68 & 87.55 & 56.27 &  51.57 & 73.47 & 43.38 &67.69 & 95.42 & 61.94 \\
        SAM2.1++~\cite{dam4sam} & 68.55 & 93.54 & 61.57 & \underline{61.99} & 91.00 & \underline{59.66} & 59.08 & 87.95 & 56.53 &\underline{53.36} & \textbf{76.36} & \underline{45.18} & \underline{68.25} & \underline{98.29} & \underline{62.42} \\
        SeC ~\cite{sec}& 69.32 & 91.89 & 62.02 & 60.73 & 88.19 & 58.33 & 57.15 & 83.85 & 54.58 & 47.86 & 66.36 & 39.53 & 68.20 & 94.35 & 62.39 \\
        SAMITE ~\cite{samite}& \underline{70.00} & \underline{94.43} & \underline{62.96}  & 61.89 & \underline{92.28} & 59.65 & \underline{59.57} & \textbf{89.67} & \underline{57.11} & 53.33 & \underline{76.20} & 45.17 & 68.04 & 95.65 & 62.27  \\
        HiM2SAM ~\cite{him2sam}& 69.48 & 93.54 & 62.51 & 61.95 & 91.37 & 59.65 & 59.46 & 89.03 & 57.04 & 50.91 & 73.03 & 43.50 & 67.73 & 95.56 & 61.98 \\
        \hline
        \rowcolor{gray!20} 
        SAMOSA (Ours) & \textbf{71.39} & \textbf{94.87} & \textbf{64.26} & \textbf {63.02} & \textbf{92.29} & \textbf{60.86} & \textbf{60.60} & 89.63 & \textbf{58.12} & \textbf{53.66} & 75.89 & \textbf{45.58} & \textbf{69.91} & \textbf{98.45} & \textbf{64.05} \\
        
        \bottomrule
    \end{tabular}
    }
    \label{tab2}
\end{table*}

\subsection{Main Results}
\label{MainRes}

Table~\ref{tab1} presents the comparative results on the test sets of $\text{LaSOT}_{ext}$~\cite{lasotext}, OTB~\cite{otb}, and TrackingNet~\cite{trackingnet}. \textbf{SAMOSA} consistently outperforms other SAM\,2-based methods across all benchmarks, particularly in AUC, Succ, and $\text{P}_{norm}$, while incurring only a marginal latency overhead. SAM2Long~\cite{sam2long} achieves comparable results to \textbf{SAMOSA} on OTB~\cite{otb} and TrackingNet~\cite{trackingnet}, but at the cost of a $32\times$ increase in $\Delta\text{Latency}$, highlighting the computational efficiency of our proposed modules. Notably, the precision of \textbf{SAMOSA} on TrackingNet~\cite{trackingnet} slightly trails SAM\,2.1~\cite{sam2} and SAMURAI~\cite{samurai}, which can be attributed to the simpler and short-term nature of TrackingNet videos, where nonlinear motion rarely occurs. Nevertheless, our method still achieves superior performance in Succ and $\text{P}_{norm}$. 

For supervised VOT methods, LoRAT~\cite{lorat}, ODTrack~\cite{odtrack}, and ARTrackV2~\cite{artrackv2} achieve performance comparable to \textbf{SAMOSA} on OTB~\cite{otb} and TrackingNet~\cite{trackingnet}, but fall behind on the larger-scale and more challenging $\text{LaSOT}_{ext}$~\cite{lasotext}, where most supervised methods perform significantly worse than SAM\,2-based approaches. This gap is likely due to the distribution shift between training and evaluation data, suggesting that supervised trackers generalize less effectively to more complex scenarios than vision foundation models.

Table~\ref{tab2} reports the results on the test sets of Anti-UAV300~\cite{antiuav300} (RGB and TIR modalities separately), Anti-UAV410~\cite{antiuav410}, DUT\,Anti-UAV~\cite{dutdatset}, and the validation set of Anti-UAV600~\cite{antiuav600}. Compared with general-purpose VOT benchmarks, \textbf{SAMOSA} demonstrates more pronounced advantages on the anti-UAV benchmarks. Although SAMITE~\cite{samite} attains slightly higher precision on Anti-UAV410~\cite{antiuav410}, its Acc and AUC are notably lower. Similarly, SAM2.1++~\cite{dam4sam} surpasses \textbf{SAMOSA} in precision on Anti-UAV600~\cite{antiuav600}, but with a $19\times$ increase in $\Delta\text{Latency}$.

Overall, the results show that \textbf{SAMOSA} achieves better generalization across different datasets than supervised VOT methods. By introducing lightweight modules with minimal latency overhead, it consistently outperforms prior SAM\,2-based methods on both general-purpose and anti-UAV benchmarks, demonstrating especially remarkable advantages in complex nonlinear tracking scenarios.

\subsection{Ablation Study}
\label{AblationStudy}

\subsubsection{Module-wise Ablation Study}
We conduct a module-wise ablation to evaluate the individual contributions of MP, EDRM, and TAMB. As shown in Table~\ref{tab:abla_modules}, the combination of all three modules yields the best overall performance. TAMB contributes most on general-purpose VOT benchmarks, while MP plays a dominant role on more complex anti-UAV benchmarks. This demonstrates TAMB’s strength in handling long-term tracking, frequent occlusion, and distractor-heavy scenes, and MP’s effectiveness in modeling complex nonlinear motion. Although EDRM provides limited additional gain when both MP and TAMB are active, its effect becomes more pronounced when either module is absent, revealing its advantage in mitigating tracking errors caused by suboptimal mask selection or memory management policies. The additional computational latency mainly originates from MP, but remains within a manageable range.

\subsubsection{Component-wise Ablation Study}

To investigate the contributions of the proposed \textit{motion}, \textit{geometry}, and \textit{semantic} cues, we conduct component-wise ablations on the EDRM, MP, and TAMB modules, respectively.

\textbf{Component-wise Ablation of MP.} Table~\ref{tab:abla_components}(a) reports the results of the MP component analysis. Incorporating either \textit{geometry} or \textit{motion} cues alongside $S_{IoU}$ surpasses the baseline strategy relying solely on $S_{IoU}$. Notably, either \textit{geometry} or \textit{motion} cues independently brings greater influence on Acc and AUC than $S_{IoU}$, proving the MP’s effectiveness in nonlinear motion modeling. Combining all three components achieves the best performance, underscoring the necessity of a comprehensive approach for mask selection.

\textbf{Component-wise Ablation of EDRM.} Table~\ref{tab:abla_components}(b) presents the component analysis for EDRM. The relative importance of components varies across tasks. On $\text{LaSOT}_{ext}$~\cite{lasotext}, \textit{semantics} contribute more than \textit{geometry}, as the general-purpose VOT task involves large, semantically rich, and often deformable targets. In contrast, on Anti-UAV300~\cite{antiuav300}, \textit{geometry} cues almost entirely dominate, since drone targets are small, semantically sparse, but geometrically stable.

\textbf{Component-wise Ablation of TAMB.} The ablation results for TAMB are shown in Table~\ref{tab:abla_components}(c). All three components contribute to filtering and evaluating memory frames, and the strategy combining all of them achieves the best results. Even when using only SAM\,2’s $S_{IoU}$ and $S_{obj}$ without additional cues, the proposed memory management mechanism brings notable improvement. The inclusion of \textit{motion} cues further enhances performance for complex scenarios.

\subsubsection{Sensitivity Analysis of Parameters}
We performed sensitivity analysis of some important parameters on Anti-UAV300 RGB. The results in Table~\ref{tab:abla_sensitivity} indicate that our method is generally robust to parameter variations and maintains stable performance under different configurations.


For MP, moderate changes in $\alpha$, $\beta$, and $\gamma$ lead to only marginal performance variations, indicating that the motion predictor remains stable under different weight configurations. This suggests that the effectiveness of MP does not rely on precise parameter tuning, and its motion prior can generalize well across different settings without significant performance degradation. 
For EDRM, varying the thresholds results in minimal performance differences. Since these thresholds mainly control the decision boundary for error detection and recovery, the consistent results indicate that the proposed mechanism can reliably identify tracking failures across a wide range of settings without requiring careful adjustment. 
For TAMB, different memory sizes and threshold values also produce consistent results. Despite changes in memory capacity and selection criteria, the performance remains stable, suggesting that the memory management strategy is robust and not sensitive to specific parameter choices.

\begin{table*}[t]
    \centering
    \caption{Ablation on our proposed modules. All $\Delta\text{Latency}$ values are tested on $\text{LaSOT}_{ext}$ and reported relative to SAM\,2.1.}
    \resizebox{\linewidth}{!}{
    \begin{tabular}{p{2.3cm}>{\centering}p{1cm}>{\centering}p{1cm}>{\centering}p{1cm}>{\centering}p{1cm}>{\centering}p{1cm}>{\centering}p{1cm}>{\centering}p{1cm}>{\centering}p{1cm}cc}
        \toprule
        \multirow{2}{*}{Methods} & \multicolumn{3}{c}{Modules} & \multicolumn{3}{c}{$\text{LaSOT}_{ext}$ (\%)} & \multicolumn{3}{c}{Anti-UAV300 RGB (\%)} & \multirow{2}{*}{$\Delta\text{Latency}$ (ms)}\\ 
        \cmidrule(lr){2-4} \cmidrule(lr){5-7} \cmidrule(lr){8-10}
        &MP & EDRM & TAMB &${AUC}$ & ${P}$ & ${P_{norm}}$ & ${Acc}$ & ${P}$ & ${AUC}$ \\
        \midrule
        SAM\,2.1~\cite{sam2} &$\times$ & $\times$ & $\times$ & 58.19 & 68.08 & 70.57 & 67.18 & 89.73 & 60.05 &  0\\
        \midrule
        \multirow{7}{*}{SAMOSA} &$\checkmark$ & $\times$ & $\times$ & 60.90 & 71.80 & 73.42 & 70.52 & 93.41 & 63.30 & +5.5 \\
        &$\times$ & $\checkmark$ & $\times$ & 60.52 & 71.43 & 73.23 & 69.49 & 93.32 & 62.37 & +3.1  \\
        &$\times$ & $\times$ & $\checkmark$ & 62.21 & 73.45 & 75.25 & 70.29 & 94.60 & 63.20 & +2.1 \\
        &$\times$ & $\checkmark$ & $\checkmark$ & 62.64 & 73.97 & 75.72 & 70.34 & 94.58 & 63.19 & +5.2 \\
        &$\checkmark$ & $\times$ & $\checkmark$ & 62.59 & 73.72 & 75.49 & 71.35 & 94.81 & 64.22 & +7.6 \\
        &$\checkmark$ & $\checkmark$ & $\times$ & 61.21 & 72.25 & 73.79 & 70.55 & 93.45 & 63.32 & +8.5 \\
        \rowcolor{gray!20} 
        &$\checkmark$ & $\checkmark$ & $\checkmark$ & \textbf{62.97} & \textbf{74.20} & \textbf{75.96} & \textbf{71.39} & \textbf{94.87} & \textbf{64.26} & +10.7 \\
        \bottomrule
    \end{tabular}
    }
    \label{tab:abla_modules}
\end{table*}

\begin{table}[t]
    \centering
    \caption{Ablation on components of MP, EDRM and TAMB.}
    \scalebox{0.89}{
    \setlength{\tabcolsep}{4pt}
    \begin{tabular}{ccccccccc}
        \toprule
        \multicolumn{3}{c}{Components} &
        \multicolumn{3}{c}{$\text{LaSOT}_{ext}$ (\%)} &
        \multicolumn{3}{c}{Anti-UAV300 RGB (\%)} \\
        \midrule
        \rowcolor{gray!20} 
        \multicolumn{9}{l}{(a) Components of MP} \\
        $S_{IoU}$ & \textit{Geometry} & \textit{Motion} & ${AUC}$ & ${P}$ & ${P_{norm}}$ & ${Acc}$ & ${P}$ & ${AUC}$ \\
        \cmidrule(lr){1-3} \cmidrule(lr){4-6} \cmidrule(lr){7-9}
        $\checkmark$ & $\times$ & $\times$ & 62.64 & 73.97 & 75.72 & 70.34 & 94.58 & 63.19\\
        $\checkmark$ & $\times$ & $\checkmark$ & 62.84 & 73.91 & 75.67 & 70.76 & 94.65 & 63.67\\
        $\checkmark$ & $\checkmark$ & $\times$ & 62.67 & 73.80 & 75.49 & 70.94 & 94.58 & 63.78\\
        $\times$ & $\checkmark$ & $\checkmark$ & 61.11 & 71.63 & 72.65 & 71.04 & 93.83 & 63.91 \\
        $\checkmark$ & $\checkmark$ & $\checkmark$ & \textbf{62.97} & \textbf{74.20} & \textbf{75.96}& \textbf{71.39} & \textbf{94.87} & \textbf{64.26}\\
        \midrule
        \rowcolor{gray!20} 
        \multicolumn{9}{l}{(b) Components of EDRM} \\
        \CompTwo{\textit{Semantics}}{\textit{Geometry}} & ${AUC}$ & ${P}$ & ${P_{norm}}$ & ${Acc}$ & ${P}$ & ${AUC}$ \\
        \cmidrule(lr){1-3} \cmidrule(lr){4-6} \cmidrule(lr){7-9}
        \CompTwo{$\times$}{$\times$} & 62.59 & 73.72 & 75.49 & 71.35 & 94.81 & 64.22\\
        \CompTwo{$\checkmark$}{$\times$} & \textbf{62.97} & 74.09 & 75.89 & 71.35 & 94.81 & 64.22\\
        \CompTwo{$\times$}{$\checkmark$} & 62.83 & 74.05 & 75.84 & 71.38 & 94.86 & 64.25\\
        \CompTwo{$\checkmark$}{$\checkmark$} & \textbf{62.97} & \textbf{74.20} & \textbf{75.96} & \textbf{71.39} & \textbf{94.87} & \textbf{64.26}\\
        \midrule
        \rowcolor{gray!20} 
        \multicolumn{9}{l}{(c) Components of TAMB} \\
        $S_{IoU}$ & $S_{obj}$ & \textit{Motion} & ${AUC}$ & ${P}$ & ${P_{norm}}$ & ${Acc}$ & ${P}$ & ${AUC}$ \\
        \cmidrule(lr){1-3} \cmidrule(lr){4-6} \cmidrule(lr){7-9}
        $\times$ & $\times$ & $\times$ & 61.21 & 72.25 & 73.79 & 70.55 & 93.45 & 63.32\\
        $\checkmark$ & $\times$ & $\times$ & 62.21 & 73.33 & 75.07 & 71.00 & 94.50 & 63.95\\
        $\times$ & $\times$ & $\checkmark$ & 61.94 & 73.06 & 74.70 & 70.42 & 93.92 & 63.74\\
        $\checkmark$ & $\checkmark$ & $\times$ & 62.45 & 73.76 & 75.37 & 71.15 & 94.65 & 64.03\\
        $\checkmark$ & $\times$ & $\checkmark$ & 62.09 & 73.05 & 74.79 &71.28 & 94.80 & 64.23\\
        $\checkmark$ & $\checkmark$ & $\checkmark$ & \textbf{62.97} & \textbf{74.20} & \textbf{75.96} & \textbf{71.39} & \textbf{94.87} & \textbf{64.26}\\
        \bottomrule
    \end{tabular}
    }
    \label{tab:abla_components}
\end{table}

\begin{table}[t]
    \centering
    \caption{Sensitivity analysis on parameters of MP, EDRM and TAMB.}
    \scalebox{0.89}{
    \setlength{\tabcolsep}{8pt}
    \begin{tabular}{C{0.08\linewidth}C{0.08\linewidth}
                C{0.08\linewidth}C{0.08\linewidth}
                ccc}
        \toprule
        \multicolumn{4}{c}{Parameters} &
        \multicolumn{3}{c}{Anti-UAV300 RGB (\%)} \\
        \midrule
        \rowcolor{gray!20} 
        \multicolumn{7}{l}{(a) Parameters of MP} \\
        $\alpha$ & $\mkern-16mu \beta_{\text{AR}}$ & $\beta_{\text{Area}}$ & $\gamma$ & ${Acc}$ & ${P}$ & ${AUC}$ \\
        \cmidrule(lr){1-4} \cmidrule(lr){5-7}
        0.85 & 0.15 & 0.10 & 0.15 & 71.39 & 94.87 & 64.26 \\
        0.75 & \textcolor{gray}{0.15} & \textcolor{gray}{0.10} & \textcolor{gray}{0.15} & 71.30 & 94.54 & 64.16 \\
        0.95 & \textcolor{gray}{0.15} & \textcolor{gray}{0.10} & \textcolor{gray}{0.15} & 71.26 & 94.70 & 64.14 \\
        \textcolor{gray}{0.85} & 0.05& 0.05 & \textcolor{gray}{0.15} & 70.91 & 94.39 & 63.78 \\
        \textcolor{gray}{0.85} & 0.25 & 0.20 & \textcolor{gray}{0.15} & 71.38 & 94.39 & 64.21 \\
        \textcolor{gray}{0.85} & \textcolor{gray}{0.15} & \textcolor{gray}{0.10} & 0.05 & 71.08 & 94.58 & 64.01 \\
        \textcolor{gray}{0.85} & \textcolor{gray}{0.15} & \textcolor{gray}{0.10} & 0.25 & 71.56 & 94.91 & 64.39 \\
        \midrule
        \rowcolor{gray!20} 
        \multicolumn{7}{l}{(b) Parameters of EDRM} \\
        $\sigma_s$ & $\tau_{ar}$ & $\tau_a$ & $\tau_s$ & ${Acc}$ & ${P}$ & ${AUC}$ \\
        \cmidrule(lr){1-4} \cmidrule(lr){5-7}
        0.10 & 0.40 & 0.40 & 0.60 & 71.39 & 94.87 & 64.26 \\
        0.30 & \textcolor{gray}{0.40} & \textcolor{gray}{0.40} & \textcolor{gray}{0.60} & 71.33 & 94.71 & 64.20 \\
        \textcolor{gray}{0.10} & 0.20 & \textcolor{gray}{0.40} & \textcolor{gray}{0.60} & 71.33 & 94.71 & 64.20 \\
        \textcolor{gray}{0.10} & 0.60 & \textcolor{gray}{0.40} & \textcolor{gray}{0.60} & 71.33 & 94.71 & 64.20 \\
        \textcolor{gray}{0.10} & \textcolor{gray}{0.40} & 0.20 & \textcolor{gray}{0.60} & 71.27 & 94.61 & 64.15 \\
        \textcolor{gray}{0.10} & \textcolor{gray}{0.40} & 0.60 & \textcolor{gray}{0.60} & 71.33 & 94.72 & 64.21 \\
        \textcolor{gray}{0.10} & \textcolor{gray}{0.40} & \textcolor{gray}{0.40} & 0.40 & 71.33 & 94.71 & 64.20 \\
        \textcolor{gray}{0.10} & \textcolor{gray}{0.40} & \textcolor{gray}{0.40} & 0.80 & 71.35 & 94.72 & 64.22 \\
        \midrule
        \rowcolor{gray!20} 
        \multicolumn{7}{l}{(c) Parameters of TAMB} \\
        $M$ & $\mu_{IoU}$ & $\mu_{obj}$ & $\mu_{m}$ & ${Acc}$ & ${P}$ & ${AUC}$ \\
        \cmidrule(lr){1-4} \cmidrule(lr){5-7}
        30 & 0.50 & 0.50 & 0.00 & 71.39 & 94.87 & 64.26 \\
        20 & \textcolor{gray}{0.50} & \textcolor{gray}{0.50} & \textcolor{gray}{0.00} & 71.19 & 94.49 & 64.09 \\
        40 & \textcolor{gray}{0.50} & \textcolor{gray}{0.50} & \textcolor{gray}{0.00} & 71.46 & 94.79 & 64.25 \\
        \textcolor{gray}{30} & 0.30 & \textcolor{gray}{0.50} & \textcolor{gray}{0.00} & 71.15 & 94.39 & 64.03 \\
        \textcolor{gray}{30} & 0.70 & \textcolor{gray}{0.50} & \textcolor{gray}{0.00} & 71.23 & 94.46 & 64.07 \\
        \textcolor{gray}{30} & \textcolor{gray}{0.50} & 0.30 & \textcolor{gray}{0.00} & 71.33 & 94.78 & 64.20 \\
        \textcolor{gray}{30} & \textcolor{gray}{0.50} & 0.70 & \textcolor{gray}{0.00} & 71.22 & 94.43 & 64.00 \\
        \textcolor{gray}{30} & \textcolor{gray}{0.50} & \textcolor{gray}{0.50} & 0.20 & 71.37 & 94.86 & 64.29 \\
        \bottomrule
    \end{tabular}
    }
    \label{tab:abla_sensitivity}
\end{table}

\begin{table*}[t]
    \vspace{-3.5mm}
	\begin{center}
		\caption{Performance comparison of different MP variants.}
		\resizebox{\linewidth}{!}{
			\begin{tabular}{cccccccccccc}
                \toprule
                \multicolumn{5}{c}{MP Configuration} & \multicolumn{3}{c}{$\text{LaSOT}_{ext}$ (\%)} & \multicolumn{3}{c}{Anti-UAV300 RGB (\%)} & \multirow{2}{*}{$\Delta\text{Latency}$}\\ 
                \cmidrule(lr){1-5} \cmidrule(lr){6-8} \cmidrule(lr){9-11}
                Backbone & \#Param & Context Len. & Training Set & Training Loss & ${AUC}$ & ${P}$ & ${P_{norm}}$ & ${Acc}$ & ${P}$ & ${AUC}$ \\
                \midrule
                KF & 0 & $\infty$ & - & - & 62.07 & 73.04 & 75.19 & 70.48 & 94.28 & 63.39 & \textcolor{gray}{Baseline}\\
                EKF & 0 & $\infty$ & - & - & 60.03 & 70.34 & 72.43 & 70.99 & 94.54 & 63.74 & +0.7 ms\\
                MLP & 7.3K & 5 & LaSOT & CIoU Loss & 61.76 & 72.85 & 74.73  & 71.01 & 94.50 & 63.84 & +0.4 ms \\
                MLP & 7.3K  & 5 & TrackingNet & CIoU Loss & 61.62 & 72.52 & 74.43 & 71.19 & 94.47 & 64.01 & +0.4 ms  \\
                LSTM & 119.3K & 3 & LaSOT & CIoU Loss & 62.19 & 73.29 & 74.92 & 70.90 & 94.45 & 63.88 & +1.3 ms\\
                LSTM & 119.3K & 4 & LaSOT & CIoU Loss & 62.62 & 74.02 & 75.73 & 71.11 & \underline{94.73} & 63.95 & +2.3 ms\\
                \rowcolor{gray!20} 
                LSTM & 119.3K & 5 & LaSOT & CIoU Loss & \underline{62.97} & \underline{74.20} & \underline{75.96} & \textbf{71.39} & \textbf{94.87} & \underline{64.26} & +2.5 ms\\
                LSTM & 119.3K & 6 & LaSOT & CIoU Loss & \underline{62.97} & 74.15 & 75.90 & 71.07 & 94.63 & 63.89 & +2.9 ms\\
                LSTM & 119.3K & 7 & LaSOT & CIoU Loss & \textbf{62.98} & \textbf{74.52} & \textbf{76.45} & 70.71 & 94.55 & 63.57 & +4.0 ms \\
                LSTM & 119.3K & 5 & TrackingNet & CIoU Loss & 62.49 & 73.37 & 75.42 & \underline{71.43} & 94.52 & \textbf{64.29} & +2.5 ms \\
                LSTM & 119.3K & 5 & LaSOT & IoU Loss & 61.85 & 72.85 & 74.51 & 70.94 & 94.34 & 63.83 & +2.5 ms \\
                LSTM & 119.3K & 5 & LaSOT & DIoU Loss & 62.72 & 74.11 & 75.77 & 70.89 & 94.53 & 63.80 & +2.5 ms \\
                \bottomrule
            \end{tabular}
        }
		\label{tab:ablation_mp_structure}
	\end{center}
	\centering
\end{table*}

\begin{figure}[t]
	\centering
	\includegraphics[width=0.48\textwidth]{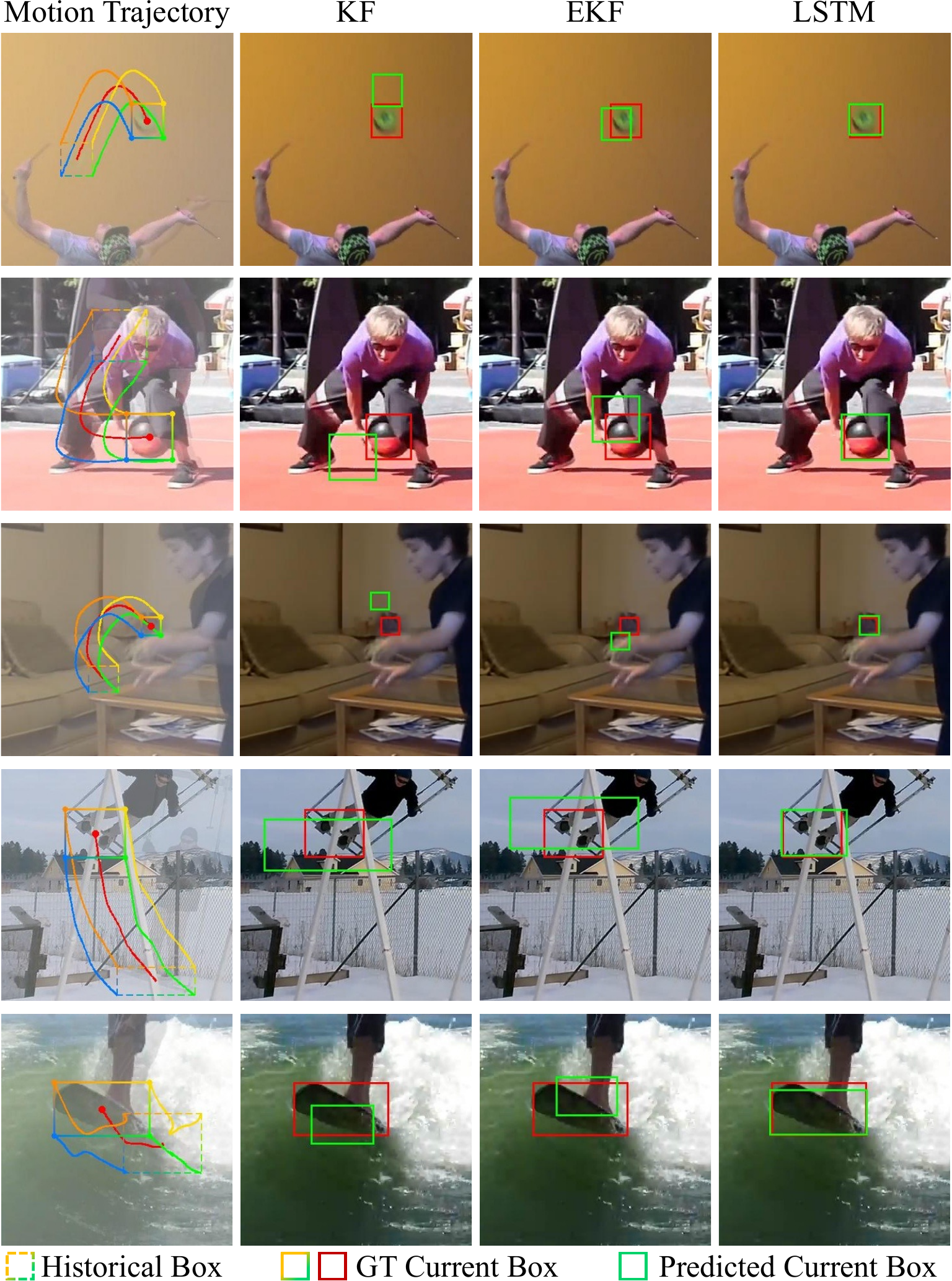}
	\caption{Visual comparison of different MP backbones under complex nonlinear motion trajectories. Frames are cropped for clarity. Zoom in for a better view.}
	\label{fig_mp_visual}
\end{figure}

\subsection{Discussion}
\label{Discussion}

In this section, we conduct additional experiments and provide further analysis of our method and its variations.

\subsubsection{Different Variations of MP}
We compare MP variants with different backbones, context lengths, training datasets, and training losses. The results are shown in Table~\ref{tab:ablation_mp_structure}.

\textbf{Backbone.} 
We replace the LSTM-based MP with a standard Kalman Filter (KF), an Extended Kalman Filter (EKF), or a lightweight MLP with only 7.3K parameters. The KF-based MP performs worse than LSTM, as it relies on linear state transitions and cannot model complex nonlinear motion effectively, as discussed in Section~\ref{subsec:mp}. Although EKF is designed for nonlinear systems, it also underperforms in our setting, even with carefully tuned noise covariance. A possible reason is that EKF depends on predefined motion models that are sensitive to noise, which is common in VOT scenarios, making it less robust than learned models such as LSTM. Replacing LSTM with the lightweight MLP further reduces latency while still achieving advancing overall performance on Anti-UAV300~\cite{antiuav300}. However, the MLP variant shows a noticeable drop on $\text{LaSOT}_{ext}$~\cite{lasotext}, where large-scale or deformable targets require a more expressive model to capture complex \textit{motion} and \textit{geometric} patterns.

Figure~\ref{fig_mp_visual} shows the predictions of KF, EKF, and LSTM on nonlinear motion trajectories. In the first three examples, when the target changes its motion direction, KF fails to adapt promptly, while EKF tends to be overly sensitive and produces aggressive predictions in response to direction changes. In the last two examples, when the target shape changes, both KF and EKF respond slowly and fail to predict the shape accurately. In contrast, LSTM, trained on trajectory annotations, can better handle abrupt changes in motion and target shape, leading to more accurate predictions under nonlinear motion.

\begin{table}[t]
    \centering
    \caption{Performance comparison across different SAM-based methods with different SAM backbones.}
    \resizebox{0.47\textwidth}{!}{
    \setlength{\tabcolsep}{2pt}
    \begin{tabular}{c>{\centering}p{1.0cm}>{\centering}p{2.3cm}>{\centering}p{0.9cm}>{\centering}p{0.9cm}>{\centering}p{0.9cm}>{\centering}p{0.9cm}>{\centering}p{0.9cm}c}
        \toprule
        \multirow{2}{*}{\raisebox{-0.7ex}{Backbone}} & \multirow{2}{*}{\raisebox{-0.7ex}{\#Param}} & \multirow{2}{*}{\raisebox{-0.7ex}{Methods}} & \multicolumn{3}{c}{$\text{LaSOT}_{ext}$ (\%)} & \multicolumn{3}{c}{Anti-UAV300 RGB (\%)} \\ 
        \cmidrule(lr){4-6} \cmidrule(lr){7-9} &&&${AUC}$ & ${P}$ & ${P_{norm}}$ & ${Acc}$ & ${P}$ & ${AUC}$ \\
        \midrule
         \multirow{3}{*}{\raisebox{-2.4ex}{SAM\,2.1-L}} & \multirow{3}{*}{\raisebox{-2.4ex}{224M}} & SAM\,2.1~\cite{sam2} & 58.2 & 68.1 & 70.6 & 67.2 & 89.7 & 60.1 \\
        &&SAMURAI ~\cite{samurai} & 61.0 & 72.1 & 73.8 & 69.3 & 92.8 & 62.3 \\
        &&SAMITE ~\cite{samite}& \underline{62.2} & \underline{73.7} & \underline{75.4} & \underline{70.0} & \underline{94.4} & \underline{63.0} \\
        &&SAMOSA (Ours) & \textbf{63.0} & \textbf{74.2} & \textbf{76.0} & \textbf{71.4} & \textbf{94.9} & \textbf{64.3} \\
        \midrule
         \multirow{3}{*}{\raisebox{-2.4ex}{SAM\,2.1-B+}} & \multirow{3}{*}{\raisebox{-2.4ex}{81M}} & SAM\,2.1~\cite{sam2} & 55.5 & 64.6 & 67.2 & 63.6 & 85.6 & 56.4 \\
        &&SAMURAI ~\cite{samurai} & 57.5 & 69.3 & 67.1 & 68.5 & 92.6 & 61.8 \\
        &&SAMITE ~\cite{samite} & \textbf{60.7} & \textbf{71.2} & \textbf{73.1} & \underline{69.0} & \textbf{94.7} & \underline{62.8} \\
        &&SAMOSA (Ours) & \underline{60.6} & \underline{71.0} & \underline{72.5} & \textbf{69.9} & \underline{94.6} & \textbf{63.5} \\
        \midrule
         \multirow{3}{*}{\raisebox{-2.4ex}{SAM\,2.1-S}} & \multirow{3}{*}{\raisebox{-2.4ex}{46M}} & SAM\,2.1~\cite{sam2} & 56.1 & 65.8 & 67.6 & 67.5 & 93.1 & 60.8 \\
        &&SAMURAI ~\cite{samurai} & 58.0 & 67.7 & 69.6 & 68.8 & 93.5 & 62.3 \\
        &&SAMITE ~\cite{samite} & \underline{59.8} & \underline{70.1} & \underline{71.7} & \underline{69.1} & \textbf{94.8} & \underline{62.8}\\
        &&SAMOSA (Ours) & \textbf{60.4} & \textbf{70.9} & \textbf{72.1} & \textbf{69.8} & \textbf{94.8} & \textbf{63.4} \\
        \midrule
         \multirow{3}{*}{\raisebox{-2.4ex}{SAM\,2.1-T}} & \multirow{3}{*}{\raisebox{-2.4ex}{39M}} & SAM\,2.1~\cite{sam2} & 52.3 & 60.3 & 62.0 & 62.9 & 84.7 & 55.7 \\
        &&SAMURAI ~\cite{samurai} & 55.1 & 63.7 & 65.6 & \underline{69.9} & 94.4 & 63.2 \\
        &&SAMITE ~\cite{samite} & \underline{57.5} & \underline{66.2} & \underline{68.0} & \underline{69.9} & \textbf{95.6} & \underline{63.7} \\
        &&SAMOSA (Ours) & \textbf{58.4} & \textbf{67.8} & \textbf{68.9} & \textbf{70.2} & 95.2 & \textbf{64.0} \\
        \midrule
        \multirow{2}{*}{SAM\,3} & \multirow{2}{*}{861M} & SAM\,3~\cite{sam3} & \underline{62.1} & \underline{73.7} & \underline{75.4} & \underline{70.9} & \underline{91.5} & \underline{64.0} \\
        &&  SAMOSA (Ours) & \textbf{65.0} & \textbf{77.0} & \textbf{78.5} & \textbf{73.8} & \textbf{95.5} & \textbf{67.2} \\
        \bottomrule
    \end{tabular}
    }
    \label{tab:different_size}
\end{table}

\textbf{Context Length.} 
For the LSTM-based MP, we vary the number of historical frames used for prediction. A context length of 5 provides a good balance between performance and latency. Longer histories brings only marginal gains on LaSOT$_{ext}$, but degrades performance on Anti-UAV300. For targets with unstable, rapidly changing motion patterns like drones, prediction benefits primarily from recent observations, while longer contexts may introduce additional noise.

\textbf{Training Set.} 
We train both LSTM-based and MLP-based MPs using bounding-box trajectories from either LaSOT~\cite{lasot}, which contains more challenging scenes, or TrackingNet~\cite{trackingnet}, which is relatively less challenging. MPs trained on either dataset achieve comparable results and consistently outperform prior methods, even when trained on the simpler TrackingNet. This indicates that MP is not heavily dependent on a specific training dataset and generalizes well across different scenarios.

\textbf{Training Loss.} 
We train MP with different loss functions, including IoU loss~\cite{iou_loss}, DIoU loss~\cite{ciou_loss}, and CIoU loss~\cite{ciou_loss}. Among them, CIoU achieves the best results, likely because it jointly considers overlap, center distance, and aspect ratio, leading to better bounding box alignment.

\begin{figure*}[t]
	\centering	\includegraphics[width=1\textwidth]{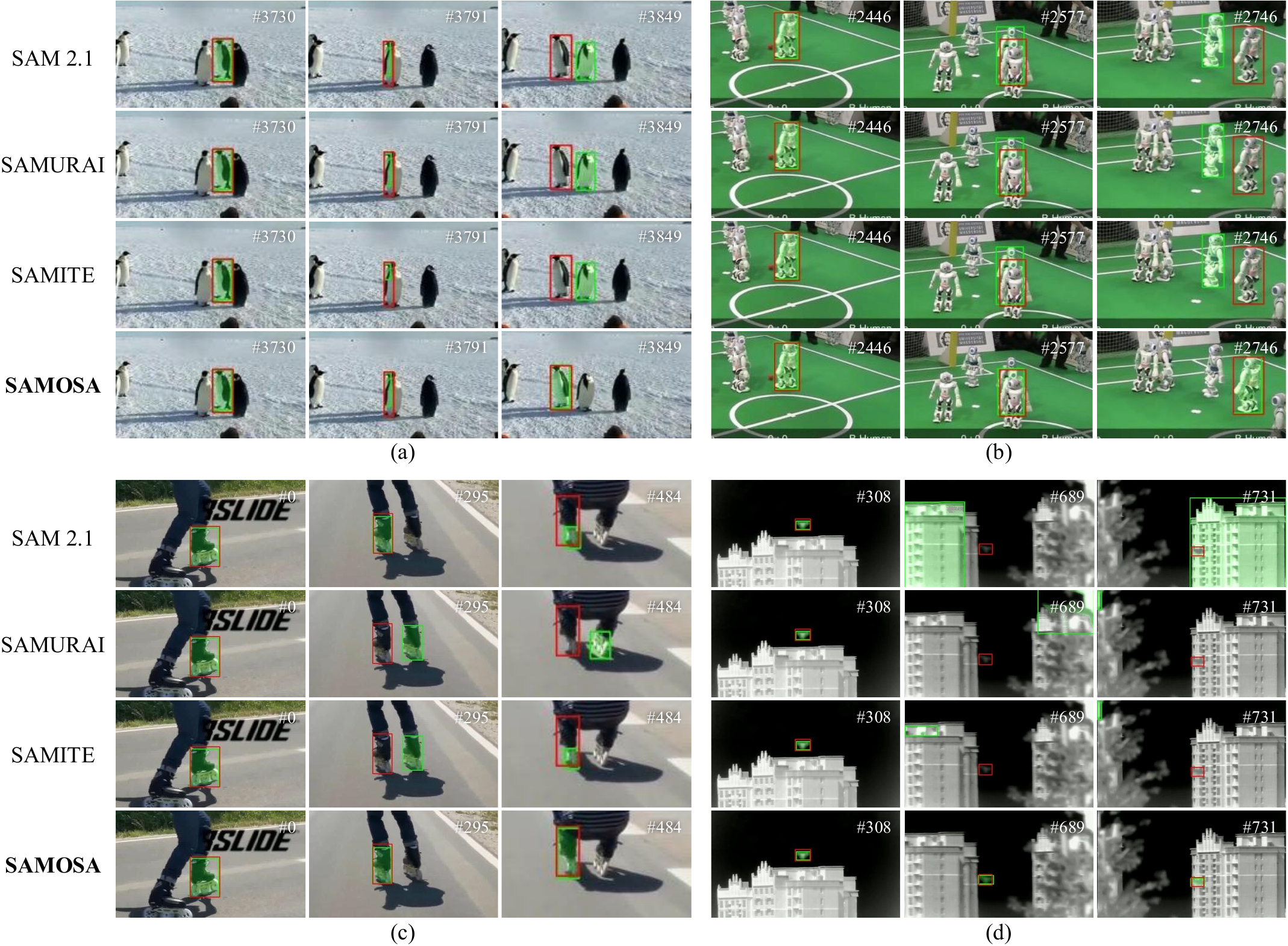}
	\caption{
     Visual comparison results. Ground truth bounding boxes are marked in red. Masks and bounding boxes predicted by methods are marked in green. Frames are cropped for clarity. Zoom in for a better view.}
	\label{fig:visual}
\end{figure*} 

\begin{table}[t]
    \centering
    \caption{Thresholds of metrics for nonlinearity analysis.}
     \scalebox{0.9}{
    \renewcommand{\arraystretch}{1.2}
    \begin{tabular}{p{5cm}c}
        \toprule
        Metric & Threshold \\
        \midrule
        Acceleration Magnitude & $20\ pixel/frame^2$ \\
        Acceleration Angle Deviation & $3\ rad/frame$ \\
        Jerk & $20\ pixel/frame^3$ \\
        \bottomrule
    \end{tabular}
    }
    \label{tab:nonlinear_thresholds}
\end{table}

\begin{table*}[t]
    \centering
    \caption{Performance comparison across different SAM\,2-based methods on linear and nonlinear split of datasets.}
    \resizebox{\textwidth}{!}{
    \begin{tabular}{lc>{\centering}p{3.8cm}>{\centering}p{1.1cm}>{\centering}p{1.1cm}>{\centering}p{1.1cm}>{\centering}p{1.1cm}>{\centering}p{1.1cm}c}
        \toprule
        \multirow{2}{*}{\raisebox{-0.7ex}{Test Sets}} & \multirow{2}{*}{\raisebox{-2.8ex}{\shortstack{Percentage of\\ Nonlinear Videos}}} & \multirow{2}{*}{\raisebox{-0.7ex}{Methods}} & \multicolumn{3}{c}{Linear Videos} & \multicolumn{3}{c}{Nonlinear Videos}\\
        \cmidrule(lr){4-6} \cmidrule(lr){7-9}
         &&& ${Acc}$ & ${P}$ & ${AUC}$ & ${Acc}$ & ${P}$ & ${AUC}$ \\
        \midrule
         \multirow{3}{*}{\raisebox{-2.4ex}{Anti-UAV300 RGB}} & \multirow{3}{*}{\raisebox{-2.4ex}{7.7\%}} & SAM\,2.1~\cite{sam2} & 65.3 & 91.2 & 58.3 & 67.4 & 90.4 & 58.9  \\
        &&SAMURAI ~\cite{samurai}& 69.4 & 93.1 & 62.5 & 68.0 & 89.2 & 59.6  \\
        &&SAMITE ~\cite{samite} & \underline{69.8} & \underline{94.4} & \underline{62.9} & \underline{71.9} & \underline{94.7} & \underline{63.3} \\
        \rowcolor{gray!20} 
        &&SAMOSA (Ours) & \textbf{71.2} & \textbf{94.8} & \textbf{64.2} & \textbf{74.0} & \textbf{95.9} & \textbf{65.5}  \\
        \midrule
         \multirow{3}{*}{\raisebox{-2.4ex}{Anti-UAV300 TIR}} & \multirow{3}{*}{\raisebox{-2.4ex}{7.7\%}}  & SAM\,2.1~\cite{sam2} & 58.4 & 86.0 & 56.3 & 70.5 & 97.7 & 66.2  \\
        &&SAMURAI ~\cite{samurai}& 60.6 & 89.7 & 58.5 & \underline{72.1} & \underline{99.8} & \underline{67.8}  \\
        &&SAMITE ~\cite{samite} & \underline{61.1} & \textbf{91.7} & \underline{59.0} & 71.2 & \textbf{99.9} & 67.0 \\
        \rowcolor{gray!20} 
        &&SAMOSA (Ours) & \textbf{62.2} & \textbf{91.7} & \textbf{60.2} & \textbf{72.9} & 99.6 & \textbf{68.5}  \\
        \midrule
         \multirow{3}{*}{\raisebox{-2.4ex}{DUT AntiUAV}} & \multirow{3}{*}{\raisebox{-2.4ex}{5.0\%}}  & SAM\,2.1~\cite{sam2} & 65.7 & 91.2 & 59.6 & 27.8 & 54.7 & 27.8  \\
        &&SAMURAI ~\cite{samurai}& 68.6 & 95.2 & 62.5 & 50.9 & \textbf{99.0} & 50.9  \\
        &&SAMITE ~\cite{samite} & \underline{68.9} & \underline{95.5} & \underline{62.9} & \underline{51.1} & \textbf{99.0} & \underline{51.1}\\
        \rowcolor{gray!20} 
        &&SAMOSA (Ours) & \textbf{70.9} & \textbf{98.5} & \textbf{64.7} & \textbf{51.4} & 98.6 & \textbf{51.4}  \\
        
        \bottomrule
    \end{tabular}
    }
    \label{tab:nonlinear_comparison}
\end{table*}

\subsubsection{Different SAM Backbones} 
We evaluate our method and baselines with different SAM backbones, including SAM\,3~\cite{sam3} and various sizes of SAM\,2.1~\cite{sam2}, on $\text{LaSOT}_{ext}$~\cite{lasotext} and Anti-UAV300~\cite{antiuav300}. The results are reported in Table~\ref{tab:different_size}. Notably, our method achieves consistently better performance across all backbone configurations. With the SAM\,2.1-B+ backbone, SAMITE~\cite{samite} achieves performance close to ours, while under other configurations it consistently underperforms our method. SAM\,3~\cite{sam3} provides clear improvements over SAM\,2.1~\cite{sam2}, and \textbf{SAMOSA} further improves its performance across both datasets.

\subsubsection{Performance in Nonlinear Scenes}
\label{sec:linearnonlinear}
To evaluate tracking performance under different motion dynamics, we quantify the motion nonlinearity of annotated bounding-box trajectories in Anti-UAV300~\cite{antiuav300} and DUT\,Anti-UAV~\cite{dutdatset}. For each frame, we compute acceleration using the second-order difference of box coordinates, and measure nonlinearity using three inter-frame indicators: acceleration magnitude, acceleration angle deviation, and jerk. A frame is labeled as \textit{nonlinear} if any of the indicators exceeds the corresponding threshold defined in Table~\ref{tab:nonlinear_thresholds}. A video is considered strongly \textit{nonlinear} when more than 45\% of its frames are classified as \textit{nonlinear}.

Based on this analysis, videos are divided into \textit{linear} and \textit{nonlinear} subsets. We evaluate our method and baselines on both splits, as shown in Table~\ref{tab:nonlinear_comparison}. Our method consistently outperforms baselines on both subsets, with particularly strong gains on the nonlinear subset of Anti-UAV300~\cite{antiuav300} RGB.

\subsection{Qualitative Results}
\label{QualitativeRes}

Figure~\ref{fig:visual} presents qualitative comparisons between our method and other SAM\,2-based trackers. In Figure~\ref{fig:visual} (a) and Figure~\ref{fig:visual} (b), our method demonstrates strong robustness against visually similar distractors. While baseline methods are easily confused by nearby objects with similar appearance, our method can maintain correct target association by jointly leveraging \textit{motion} consistency and \textit{geometric} constraints, resulting in more stable and accurate tracking. In Figure~\ref{fig:visual} (c), when two skating shoes frequently cross and occlude each other, baseline methods either drift to the wrong target or produce incomplete masks, while our method maintains robust and precise tracking under severe mutual occlusions. In Figure~\ref{fig:visual} (d), thermal infrared scenes impose a challenging inter-modality setting for SAM\,2-based approaches, due to limited semantic information, similar colors, and frequent occlusions. Even under such conditions, our method consistently tracks the target for most of the time and successfully re-acquires it after occlusion.

\section{Conclusion}
\label{Conclusion}

In this paper, we have presented \textbf{SAMOSA}, a framework that leverages \textit{motion}, \textit{geometry}, and \textit{semantic} cues to address complex nonlinear visual object tracking. Existing methods often struggle to model nonlinear motion patterns prevalent in VOT scenarios and lack explicit mechanisms for error detection and recovery. To address these limitations, we design a Motion Predictor (MP) based on high-order Markov modeling to capture nonlinear motion dynamics. We further introduce an Error Detection-Recovery Module (EDRM) to mitigate error accumulation by explicitly detecting and rectifying tracking failures. In addition, a Target-Aware Memory Bank (TAMB) is proposed to enable efficient memory management by retaining representative reference frames.

Extensive experiments demonstrate that, with controllable latency overhead, our method achieves state-of-the-art performance and strong generalization ability on both general-purpose VOT benchmarks and challenging anti-UAV tracking benchmarks, demonstrating consistent robustness and effectiveness. As a lightweight and pluggable adapter, \textbf{SAMOSA} can be seamlessly integrated into future generations of Segment Anything models. Future work will explore extending our framework to broader tasks such as multi-object tracking, referring object segmentation, and 3D object segmentation. 



\bibliographystyle{IEEEtran}
\bibliography{main}


 

\begin{IEEEbiography}[{\includegraphics[width=1in,height=1.25in,clip,keepaspectratio]{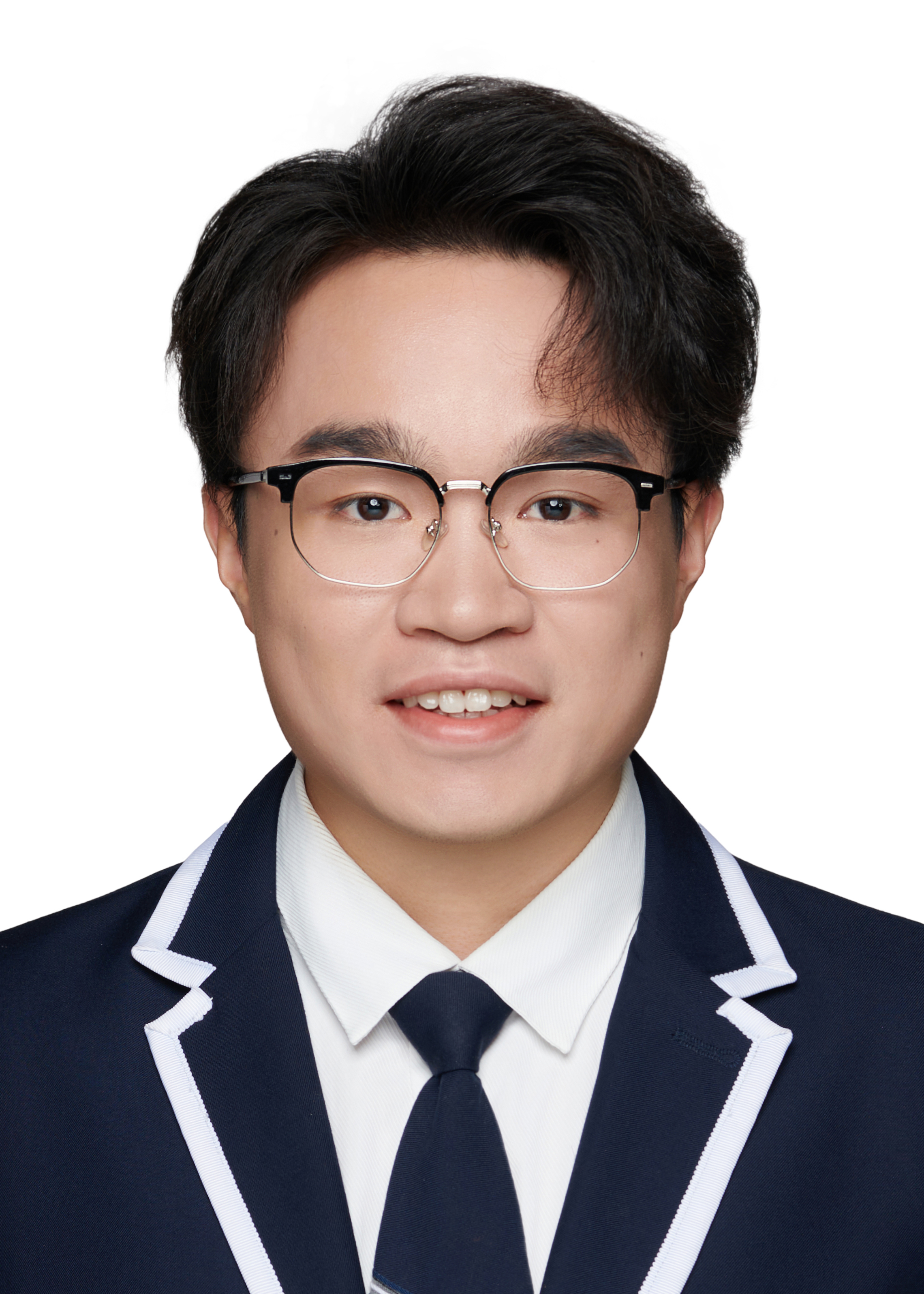}}]{Deyi Zhu} received the B.S. degree from the Department of Automation, Tsinghua University, in 2025. He is currently pursuing the Ph.D degree with Tsinghua Shenzhen International Graduate School, Tsinghua University. His current research interests include computer vision and embodied intelligence.
\end{IEEEbiography}
\begin{IEEEbiography}[{\includegraphics[width=1in,height=1.25in,clip,keepaspectratio]{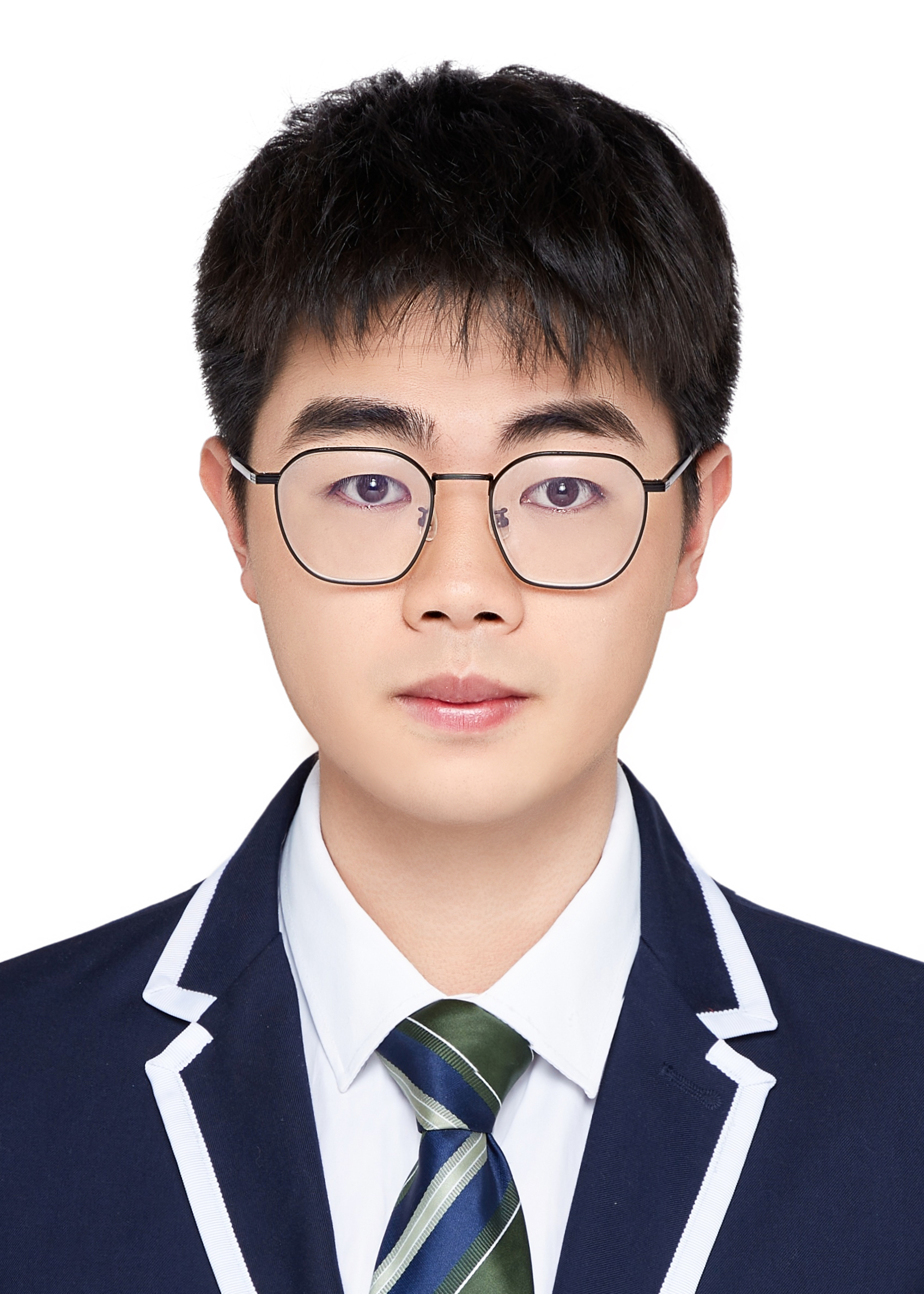}}]{Yuji Wang} received the B.S. degree in Electric and Electronic Engineering from the University of Electronic Science and Technology of China (UESTC) in 2024. He is currently a second-year master student with the Shenzhen International Graduate School, Tsinghua University, supervised by Prof. Yansong Tang. His research interests focus on computer vision, including vision-language models, tool-calling, multimodal learning, image/video segmentation and tracking. He has published papers in top conferences such as CVPR, AAAI and ECCV, and conducted research internships in multimodal learning and function calling related fields.
\end{IEEEbiography}
\begin{IEEEbiography}[{\includegraphics[width=1in,height=1.25in,clip,keepaspectratio]{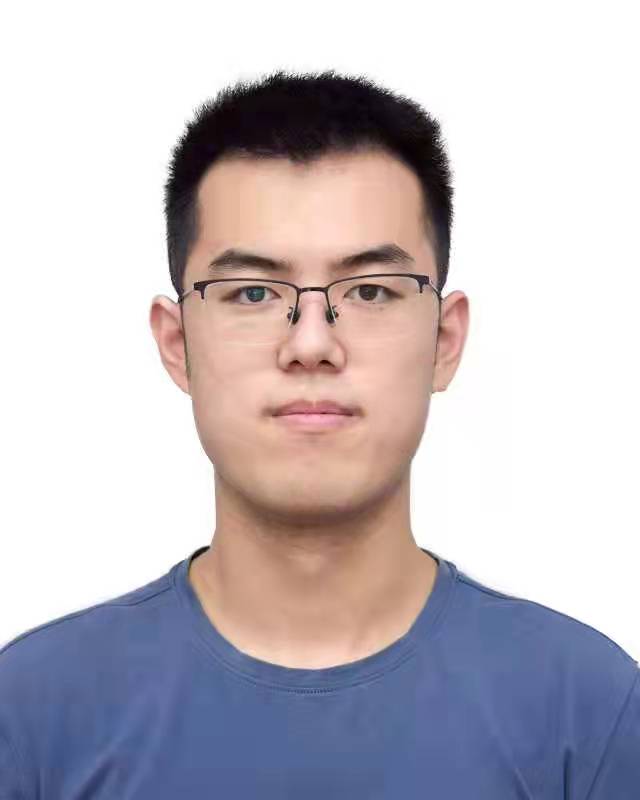}}]{Yong Liu} received the B.Eng. degree from Shandong University in 2020. He is currently pursuing the Ph.D degree with Tsinghua Shenzhen International Graduate School, Tsinghua University. His current research interests include fine-grained video understanding and multimodal understanding.
\end{IEEEbiography}
\begin{IEEEbiography}[{\includegraphics[width=1in,height=1.25in,clip,keepaspectratio]{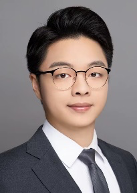}}]{Yansong Tang} (Member, IEEE) 
 received the B.S. and Ph.D. degrees from the Department of Automation, Tsinghua University, in 2015 and 2020, respectively. From 2020 to 2022, he served as a Postdoctoral Fellow at the Department of Engineering Science, University of Oxford. He is currently a tenure-track Associate Professor of Shenzhen International Graduate School, Tsinghua University. In recent years, he has authored more than 40 papers in top peer-reviewed journals and conferences such as IEEE Transactions on Pattern Analysis and Machine Intelligence, IEEE Transactions on Image Processing, and CVPR. His research interests include computer vision, pattern recognition, and video processing.
\end{IEEEbiography}
\begin{IEEEbiography}[{\includegraphics[width=1in,height=1.25in,clip,keepaspectratio]{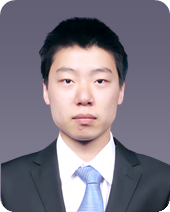}}]{Bingyao Yu} received the B.S. and Ph.D. degrees both in the Department of Automation, Tsinghua University, China, in 2018 and 2023. He is currently a postdoctoral researcher with the Department of Automation, Tsinghua University.  His current research interests include computer vision, AI security and embodied intelligence. He has published more than 10 scientific papers in  TIP, TIFS, CVPR, ICCV and ACMMM. He serves as a regular reviewer member for a number of journals and conferences, e.g. TPAMI, TIP, ICML, ICLR, NeurIPS, CVPR, ECCV, and ICCV.
\end{IEEEbiography}
\begin{IEEEbiography}[{\includegraphics[width=1in,height=1.25in,clip,keepaspectratio]{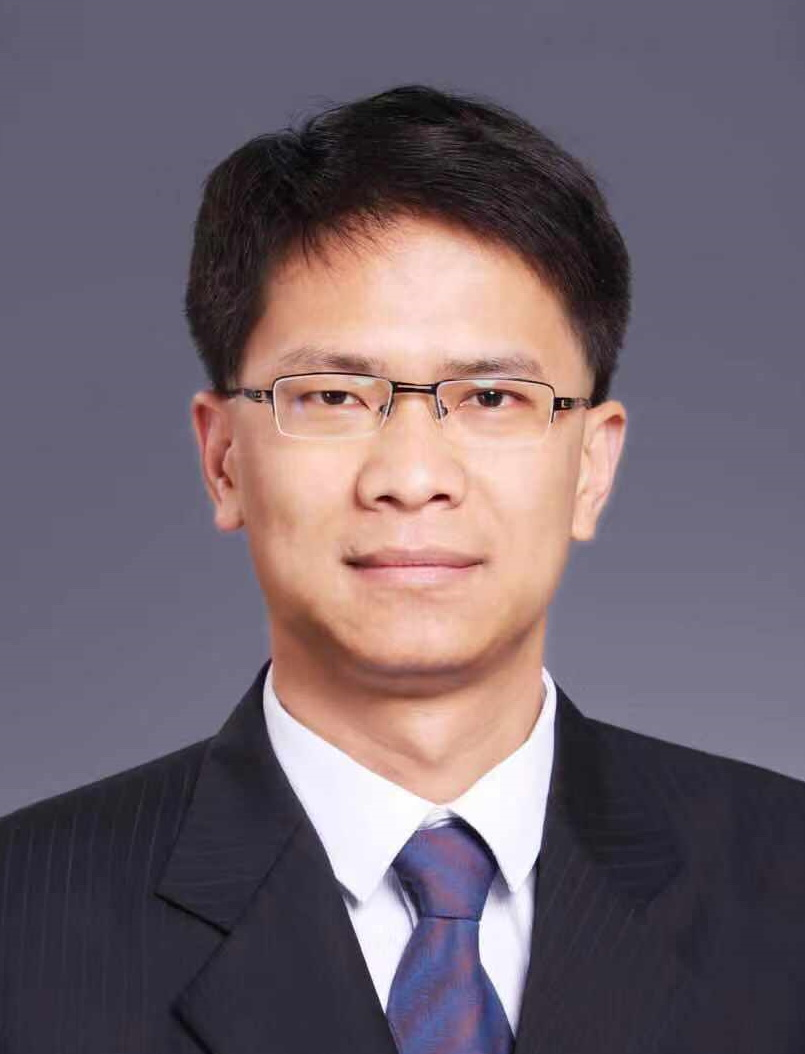}}]{Jiwen Lu} (Fellow, IEEE) received the B.Eng. degree in mechanical engineering and the M.Eng. degree in electrical engineering from the Xi’an University of Technology, Xi’an, China, in 2003 and 2006, respectively, and the Ph.D. degree in electrical engineering from Nanyang Technological University, Singapore, in 2012. From 2011 to 2015, He was with the Advanced Digital Sciences Center, Singapore. In November 2015, he joined the Department of Automation, Tsinghua University, where he is currently a full professor and the deputy chair of the department. His current research interests include computer vision, pattern recognition, multimedia computing, and intelligent robotics. He serves as the Co-Editor-of-Chief for Pattern Recognition Letters, an Associate Editor for the IEEE Transactions on Image Processing, the IEEE Transactions on Circuits and Systems for Video Technology, and the IEEE Transactions on Biometrics, Behavior, and Identity Sciences, and Pattern Recognition. He was a recipient of the National Natural Science Funds for Distinguished Young Scholar. 
He is an IEEE/IAPR Fellow.
\end{IEEEbiography}
\begin{IEEEbiography}[{\includegraphics[width=1in,height=1.25in,clip,keepaspectratio]{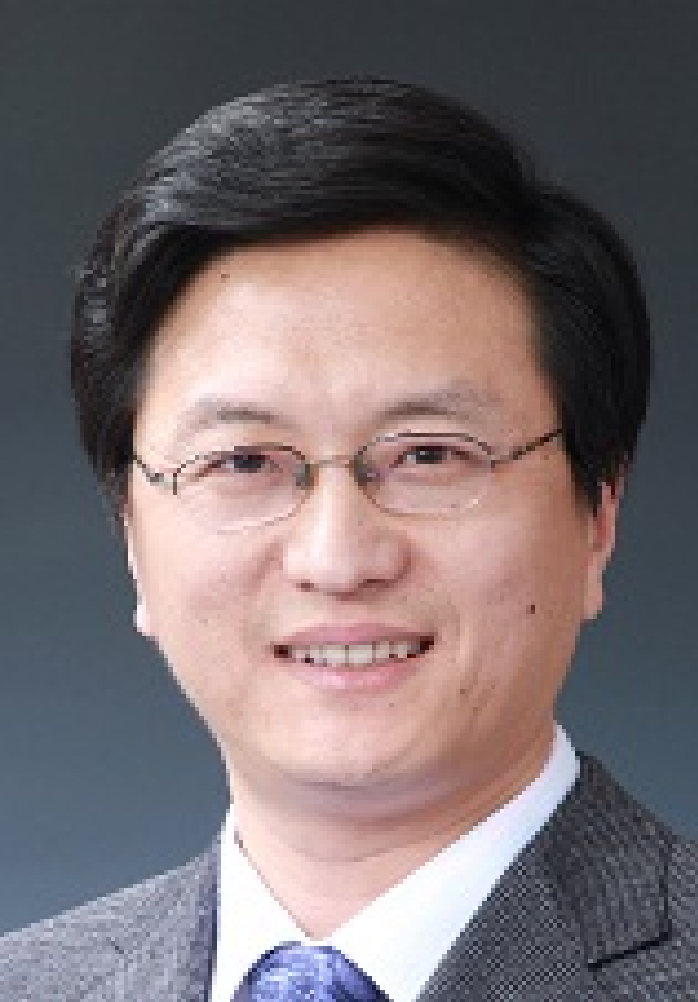}}]{Jie Zhou} (Fellow, IEEE) received the B.S. and M.S. degrees from the Department of Mathematics, Nankai University, Tianjin, China, in 1990 and 1992, respectively, and the Ph.D. degree from the Institute of Pattern Recognition and Artificial Intelligence, Huazhong University of Science and Technology, Wuhan, China, in 1995. From 1995 to 1997, he was a Postdoctoral Fellow with the Department of Automation, Tsinghua University, Beijing, China. Since 2003, he has been a Full Professor with the Department of Automation, Tsinghua University. In recent years, he has authored more than 300 papers in peer-reviewed journals and conferences. Among them, more than 100 papers have been published in top journals and conferences, such as IEEE Transactions on Pattern Analysis and Machine Intelligence, IEEE Transactions on Image Processing, and CVPR. His research interests include computer vision, pattern recognition, and image processing. He is also an Associate Editor for IEEE Transactions on Pattern Analysis and Machine Intelligence and two other journals. He was the recipient of the National Outstanding Youth Foundation of China Award. 
He is an IEEE/IAPR Fellow.
\end{IEEEbiography}



\vfill

\end{document}